%% file: V2_main.tex
\documentclass{article} 
\usepackage{amssymb}
\usepackage{multirow}
\usepackage[table]{xcolor}
\usepackage{iclr2025_conference,times}
\iclrfinalcopy

\usepackage{multirow}

\usepackage{microtype}
\usepackage{hyperref}
\usepackage{url}
\usepackage{booktabs}
\usepackage[most]{tcolorbox}
\usepackage{listings}
\usepackage{xspace}
\usepackage{cleveref}
\usepackage{fontawesome5}
\usepackage{wrapfig}
\usepackage{listingsutf8}
\usepackage{graphicx}
\usepackage{subcaption}

\usepackage{arydshln}
\usepackage{threeparttable}

\newcommand{\prover}{Goedel-Prover-SFT\xspace}
\newcommand{\provernew}{Goedel-Prover-V2\xspace}

\newcommand{\putnam}{PutnamBench\xspace}

\newcommand{\miniff}{MiniF2F\xspace}

\newcommand{\mobench}{MathOlympiadBench\xspace}

\lstdefinestyle{leanstyle}{
    language=lean,
    basicstyle=\small\ttfamily,
    keywordstyle=\color{blue},
    stringstyle=\color{red!60!black},
    commentstyle=\color{green!60!black},
    breaklines=true,
    showstringspaces=false,
    numbers=left,      
    numberstyle=\tiny\color{gray}, 
    stepnumber=1,
    numbersep=5pt,
    mathescape=true
}

\newtcblisting{leancode}[1][]{
    listing engine=listings,
    listing style=leanstyle,
    boxrule=0.5pt,
    colback=white,
    colframe=gray!70,
    arc=0pt, 
    boxsep=0pt,
    left=2pt, right=2pt, top=2pt, bottom=2pt,
    listing only,
    title style={colback=white, colframe=gray!70, coltext=black},
    fonttitle=\bfseries,
    #1, 
}

\newtcblisting{plainleancode}[1][]{
    listing engine=listings,
    listing only,
    listing options={
        language=lean,
        basicstyle=\small\ttfamily,
        keywordstyle=\color{black}, 
        stringstyle=\color{black},  
        commentstyle=\color{black}, 
        breaklines=true,
        showstringspaces=false,
        numbers=none,              
        mathescape=true
    },
    boxrule=0.5pt,
    colback=white,
    colframe=gray!70,
    arc=0pt,
    boxsep=0pt,
    left=2pt, right=2pt, top=2pt, bottom=2pt,
    breakable,
    fonttitle=\bfseries,
    #1,
}

\definecolor{commentcolor}{RGB}{0,128,0}
\definecolor{symbolcolor}{rgb}{0.5,0.2,0.8} 
\definecolor{sortcolor}{rgb}{0.5,0.5,0.8}  
\definecolor{darkblue}{rgb}{0, 0, 0.5}
\hypersetup{colorlinks=true, citecolor=darkblue, linkcolor=darkblue, urlcolor=darkblue}

\title{{\provernew: Scaling Formal Theorem Proving with Scaffolded Data Synthesis and Self-Correction}}

\author{Yong Lin$^{1}\thanks{Core Contributor.}$\ ,\ Shange Tang$^{1\ 2\ *}$,\ Bohan Lyu$^{3\ *}$,\ Ziran Yang$^{1\ *}$,\ Jui-Hui Chung$^{1\ *}$,\\
\textbf{Haoyu Zhao$^{1\ *}$,\ Lai Jiang$^{7\ *}$,\ Yihan Geng$^{8\ *}$,\ Jiawei Ge$^{1}$,\ Jingruo Sun$^{4}$,}\\
\textbf{Jiayun Wu$^{3}$,\ Jiri Gesi$^{6}\ $\thanks{This work is independent of and outside of the work at Amazon.}\ ,\ Ximing Lu$^{2}$, \ David Acuna$^{2}$,\ Kaiyu Yang$^{5}\ \thanks{All experiments and data processing were conducted outside Meta.}$\ ,}\\ \textbf{Hongzhou Lin$^{6\ *}$\footnotemark[2]\ ,
Yejin Choi$^{2\ 4}$,\ Danqi Chen$^{1}$,\ Sanjeev Arora$^{1}$,\ Chi Jin$^{1\ *}$}
\\
$^1$Princeton Language and Intelligence, Princeton University \quad
$^2$NVIDIA \\
$^3$Tsinghua University \quad
$^4$Stanford University \quad 
$^5$Meta FAIR \quad
$^6$Amazon \\
$^7$Shanghai Jiao Tong University \quad
$^8$Peking University
}

\begin{document}

\maketitle



\begin{abstract}


We introduce Goedel-Prover-V2, a series of open-source language models that set a new state-of-the-art in automated theorem proving. Built on the standard expert iteration and reinforcement learning pipeline, our approach incorporates three key innovations: (1) Scaffolded data synthesis: We generate synthetic tasks of increasing difficulty to train the model to master increasingly complex theorems; (2) Verifier-guided self-correction: We enable the model to iteratively revise its proofs by leveraging feedback from the Lean compiler; (3) Model averaging: We merge model checkpoints to mitigate the decrease in model output diversity in later stages of training. Our small model, Goedel-Prover-V2-8B, reaches 84.6\% pass@32 on MiniF2F and outperforms DeepSeek-Prover-V2-671B under the same metric, despite being 80X smaller. Our flagship model, Goedel-Prover-V2-32B, achieves 88.1\% on MiniF2F at pass@32 in standard mode and 90.4\% in self-correction mode, outperforming prior SOTA by a large margin. Additionally, our flagship model solves 86 problems on PutnamBench at pass@184, securing the first place among open-source models on the leaderboard, surpassing DeepSeek-Prover-V2-671B’s record of solving 47 problems by pass@1024 with a significantly smaller model size and compute budget. At the time of its release (July–August 2025), Goedel-Prover-V2 achieves the strongest overall performance among all open-source theorem provers. It also ranks among the top-performing models---including closed-source systems with publicly reported performance---under a constrained test-time compute budget.
Our models, code, and data are released at \url{https://github.com/Goedel-LM/Goedel-Prover-V2}.

\end{abstract}

\input{sections/intro}

\input{sections/methods}

\input{sections/experiments}

\input{sections/related}

\input{sections/conclusion}

\bibliography{iclr2025_conference}
\bibliographystyle{iclr2025_conference}

\newpage
\appendix

\input{sections/appendix}
\end{document}

%% file: sections/intro.tex
\section{Introduction}

Automated theorem proving (ATP) is a grand challenge for AI, requiring the construction of step-by-step, machine-verifiable proofs in formal languages like Lean~\citep{de2015lean,moura2021lean}. Unlike reasoning in natural language, ATP demands a completely rigorous logical flow, which poses an exceptional challenge for AI systems. The field has advanced rapidly in recent years; for example, DeepMind's AlphaProof~\citep{alphaproof} and AlphaGeometry~\citep{trinh2024solving,chervonyi2025gold} demonstrated that AI systems are capable of achieving IMO (International Math Olympiad) medal-level performance. Furthermore, open-source efforts such as DeepSeek-Prover-V2~\citep{ren2025deepseek} and Kimina‑Prover~\citep{wang2025kimina} have achieved impressive results on benchmarks like MiniF2F~\citep{zheng2021minif2f} and PutnamBench~\citep{tsoukalas2024putnambench}, demonstrating the effectiveness of leveraging the reasoning ability of LLMs through chain-of-thoughts. These successes, however, typically depend on massive models (e.g., DeepSeek-Prover-V2 has 671B parameters) or computationally intensive inference (e.g., the concurrent work Seed-Prover \citep{chen2025seedproverdeepbroadreasoning}), which involves complex search algorithms and enormous search budgets.


In this work we introduce Goedel-Prover-V2, a new series of open-source models for automated theorem proving in Lean that establish a new state-of-the-art in both performance and computational efficiency. The models are capable of generating whole proof and leveraging verifier feedback for iterative self-correction. Our flagship 32B model achieves 88.1\% pass@32 on the MiniF2F benchmark, improving to 90.4\% with self-correction. This performance surpasses both the concurrent 72B Kimina-Prover and the previous SOTA 671B DeepSeek-Prover-V2, while using significantly fewer parameters (\Cref{fig:combined-performance}). On the more challenging PutnamBench, our model solves 44 problems (57 with self-correction), more than doubling the number solved by DeepSeek-Prover-V2 under the same metric. The efficiency of our approach is further underscored by our 8B model, which alone outperforms the 671B DeepSeek-Prover-V2 on MiniF2F despite being nearly 80 times smaller. At the time of its release (July–August 2025), Goedel-Prover-V2 achieves the strongest overall performance among all open-source theorem provers. It also ranks among the top-performing models---including closed-source systems with publicly reported performance---under small test-time compute budgets.

\input{figures/figure_combined_performance}

The key to these gains lies in novel design across the framework, data, and training pipelines. We highlight the main components below:
\begin{itemize}
    \item \textbf{Verifier-guided self-correction:} Framework-wise, we incorporate feedback from the Lean compiler (verifier) into the model input to enable the error-correction ability of our theorem prover (\emph{verifier-guided self-correction}). While correcting errors based on verifier feedback has been studied in theorem proving~\citep{first2023baldur} and coding~\citep{olausson2024is,chen2024teaching,bouzenia2024repairagent}, we further incorporate this into models that generate long chain-of-thought (CoT) reasoning, which is effective for complex reasoning tasks such as ATP~\citep{jaech2024openai,guo2025deepseek,wang2025kimina,ren2025deepseek}.

    \item \textbf{Scaffolded data synthesis:} Successfully combining long CoT with verifier-guided error correction requires special efforts on curating data. In addition to formalizing existing math problems and curating data for error correction, we augment our training statement set through \emph{scaffolded data synthesis}. This technique creates math problems at an appropriate difficulty level to provide the model with better learning signals.

    \item \textbf{Model averaging:} Our training recipe extends beyond standard expert iteration and reinforcement learning. We also apply \emph{model averaging}~\citep{wortsman2022robust} to mitigate the decrease in model output diversity that can occur in the later stages of training.
\end{itemize}

Overall, our results demonstrate that the frontier of formal theorem proving can be advanced without access to extremely large models, a vast computational resources, or proprietary technology. We hope that our open-source theorem prover series, Goedel-Prover-V2, will enable the community to build upon them and accelerate progress toward AI systems that can reliably solve and verify complex mathematical problems, and ultimately bridge the long-standing divide between intuitive human reasoning and formal proof verification.

The following paper is organized as follows: in \Cref{sec:methods}, we introduce our main methods to train Goedel-Prover-V2, including the verifier-guided self-correction framework (\Cref{sec:correction}), scaffolded data synthesis (\Cref{sec:datasynthesis}), and the training pipeline (\Cref{sec:training}). In \Cref{sec:eval}, we present the evaluation results on Goedel-Prover-V2, including evaluation on both standard and our curated benchmarks~ (\Cref{sec:results}), scaling behavior under different test-time budgets~ (\Cref{sec:scaling}), and a study of reinforcement learning and model averaging~(\Cref{sec:rlavg}). Finally, we review the previous works (\Cref{sec:related}) and discuss the connection with Goedel-Prover-V2.

%% file: figures/figure_combined_performance.tex
\begin{figure}[!t]
    \centering
    \includegraphics[width=\linewidth]{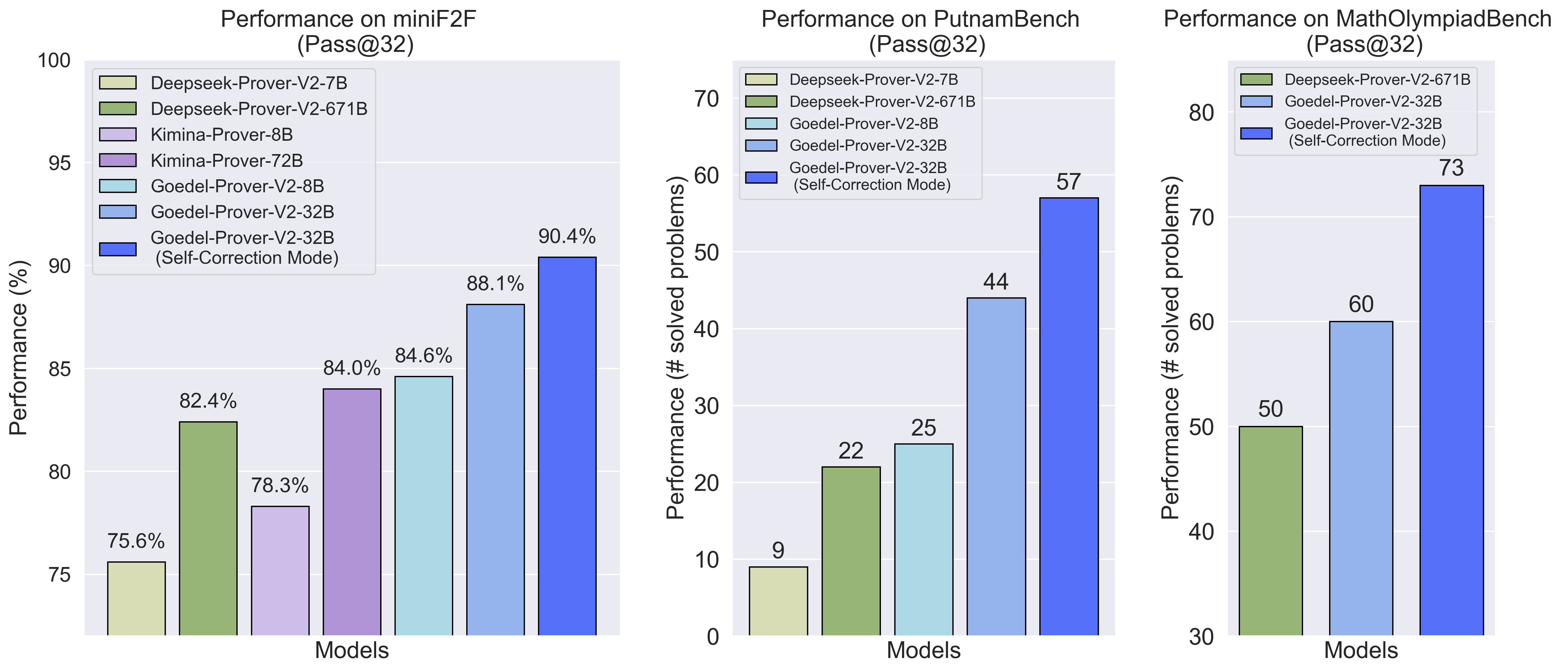}
    \caption{Performance of Goedel-Prover-V2 on different benchmarks under pass@32.}
    \label{fig:combined-performance}
\end{figure}

%% file: sections/methods.tex
\section{Method}\label{sec:methods}

In this section, we present our method in detail. We start with the key framework innovation compared to the previous vanilla whole-proof generation methods~(e.g. DeepSeek-Prover-V2) by utilizing the feedback from the Lean compiler to guide the proof-correction procedure (\Cref{sec:correction}). Then, in \Cref{sec:datasynthesis}, we demonstrate how to curate the training data (statements), with an emphasis on augmenting the data through scaffolded data synthesis. Based on the curated data, we present the details of training our theorem prover in \Cref{sec:training}, which includes supervised fine-tuning (SFT), reinforcement learning (RL), and model averaging. Section~\ref{sect:overall_pipeline} provides a summary of the overall framework.

\subsection{Chain-of-Thought and Self-Correction}\label{sec:correction}


Prevailing paradigms for whole-proof generation in automated theorem proving have largely been ``end-to-end," where a model generates a complete formal proof from a theorem statement in a single pass~\citep{xin2024deepseekv15,lin2025goedel,dong2025stp}. Recent work has significantly advanced this approach by leveraging the long-chain-of-thought reasoning capabilities of large models~\citep{wang2025kimina,ren2025deepseek}.

A key distinction between informal and formal proof generation lies in the availability and utilization of compilation (verification) feedback from proof assistants like Lean or Coq. Humans naturally leverage such feedback to iteratively revise their proofs. Recent works have shown that integrating the verifier's error messages or tactic verification outcomes into proof generation significantly improves synthesis accuracy~\citep{first2023baldur}.

Our framework formalizes this intuition by explicitly incorporating verifier feedback within the whole-proof generation loop. We structure the pipeline so that, after an initial proof attempt, verification failures are parsed and communicated back into the model as corrective guidance. The model then generates proof repairs, leading to an iterative self‑correction process.

\subsection{Curating Formal Statements}\label{sec:datasynthesis}


In this part, we detail our methods for curating a large, high-quality dataset of Lean 4 statements, which is essential for expert iteration and RL in formal theorem proving. We start with the formalizer training, which is the core of translating existing math problems written in natural language into Lean statements. We then present our scaffolded data synthesis pipeline that creates math problems at an appropriate difficulty level, which includes a lightweight formal-based method that utilizes the Lean system, and an informal-based system for large-scale data augmentation.


\begin{wraptable}{r}{0.45\textwidth}
\vspace{-5mm}
\centering
\begin{tabular}{lcc}
\toprule
\textbf{Name} &  \textbf{Pass} &  \textbf{Failed} \\
\midrule
Kimina-autoformalizer & 161 & 139 \\
Goedel-Formalizer-V2  & 228 & 72 \\
\bottomrule
\end{tabular}
\caption{Comparison of different formalizers on 300 Omni-math problems.}
\label{tab:formalizer}
\vspace{-5mm}
\end{wraptable}



\paragraph{Formalizer Training.}

Existing Lean datasets (e.g., Goedel-Pset-v1) contain many low-quality problems—human evaluation on a sampled subset shows that over 80\% of unsolved problems are incorrectly formalized. This highlights the need for stronger formalizers. We train our formalizer using the standard expert iteration pipeline, which combines Lean syntax checks and semantic evaluation via LLMs to assess translation quality. Only statements that pass both checks are included in the next iteration of SFT. Details of the LLM semantic evaluation prompt are provided in Appendix~\ref{app:prompt}.

To initialize expert iteration, we prompt Claude Sonnet 4 to generate 50K formalized statements, along with corresponding reasoning traces. Incorporating reasoning capability enables our formalizer to outperform previous models such as Goedel-Formalizer-V1 and Kimina-Autoformalizer, which lack this feature.

We evaluate our model Goedel-Formalizer-V2 and Kimina-Autoformalizer on a benchmark of 300 OmniMath problems. Results are shown in Table~\ref{tab:formalizer}.

\begin{figure}[!t]
    \centering
    \includegraphics[width=0.8\linewidth]{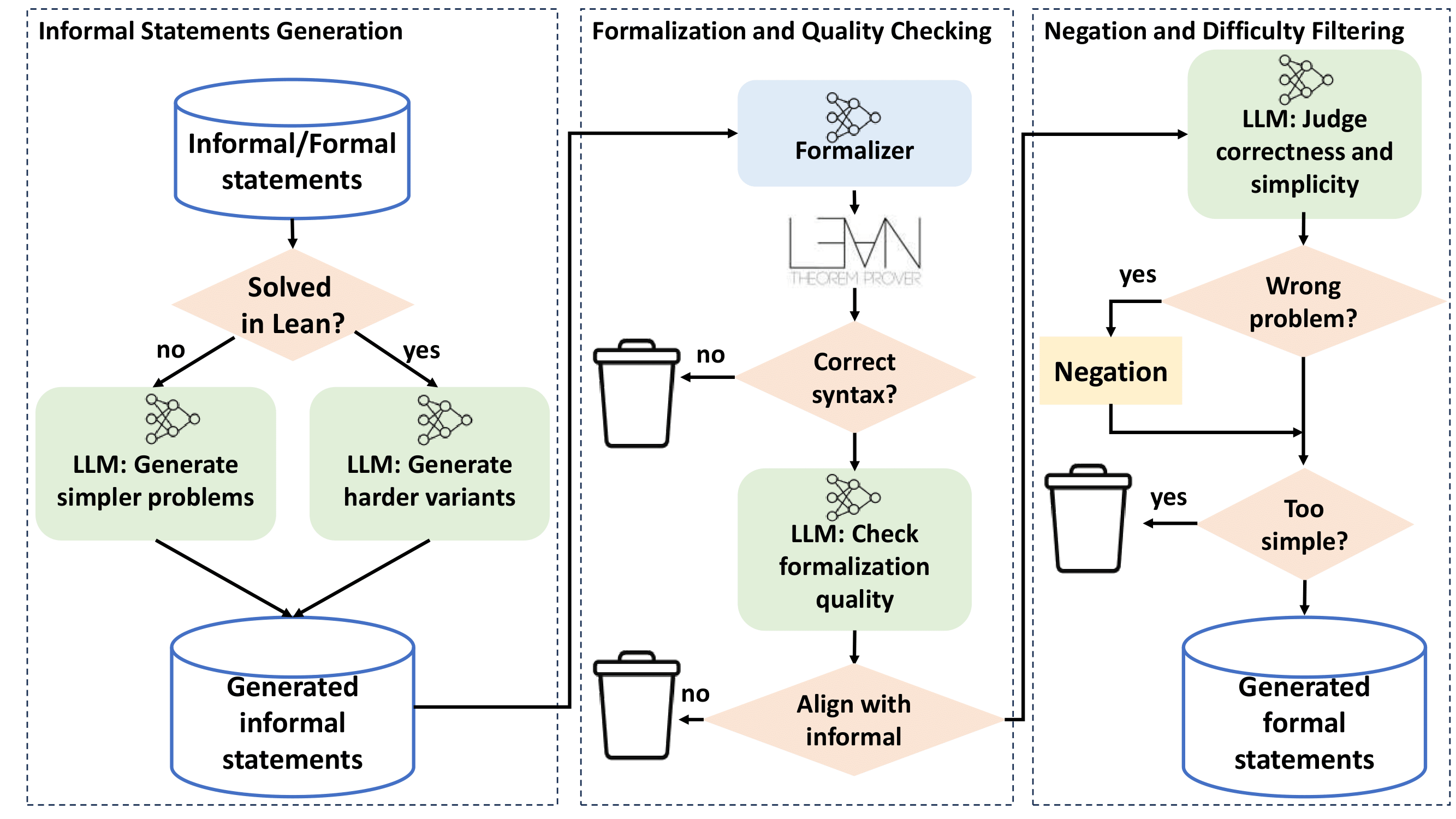}
    \caption{Our \textbf{informal-based scaffolded data synthesis pipeline} with three parts: (1) informal statement generation; (2) formalization and quality checking; and (3) negation and difficulty filtering.}
    \label{fig:data-synthesis}
\end{figure}

\paragraph{Formal-based scaffolded data synthesis.}

When the prover fails to find a proof for a challenging problem, we can still leverage the proof attempt to generate simpler, related problems. The intuition is that even an incorrect proof attempt may introduce valid subgoals that represent easier subproblems. We utilize a powerful tactic in Lean, $\texttt{extract\_goal}$, to capture the unsolved states of a proof. These extracted goals (together with the preconditions), which are well-formed mathematical statements, are then used to augment our training data for the next phase. Since an extracted statement is not guaranteed to be provable, we also include its negation, thereby training the model to recognize both true and false propositions.

\paragraph{Informal-based scaffolded data synthesis.}

Another way of scaffolded data synthesis is to leverage the current LLMs' mathematical reasoning ability in natural language, to create math statements at the right level for the model to learn. For a given problem, we prompt an LLM (Qwen3-32B) to generate simpler/sub-problems if it is unsolved, or harder variants if it is already solved. To improve generation quality, we first have the LLM attempt a natural language solution to the original problem, using its output as context. The generated informal statements are then formalized into Lean by the formalization pipeline we described in \textbf{Formalizer Training}. To avoid the inference overhead of incorrect or trivial statements, we use an LLM-based filter that assesses each statement for correctness and difficulty, where trivial or mathematically incorrect statements are discarded (the negation of incorrect statements are also added to the dataset). This filtering process significantly accelerates data synthesis, with a minor trade-off in potentially discarding some valid statements due to LLM judgment errors. The prompts and further details are in \Cref{sec:informal-scaffolded}.



\subsection{Training Algorithms}\label{sec:training}

\paragraph{Supervised fine-tuning and expert iteration.}
We follow the standard pipeline of Expert Iteration by iterating between using the model to conduct large-scale inference on the statement sets, collecting correct proofs with reasoning traces, and fine-tuning the model further by Supervised Fine-tuning (SFT) on the collected samples.


\paragraph{Reinforcement learning.}

We aim to train a model capable of generating complete proofs and correcting its own errors via verifier-guided self-correction. To this end, we adopt an efficient hybrid strategy. We first sample multiple complete proofs in parallel. For each sampled proof, we perform serial self-correction, generating one correction per round. This serial correction process aligns well with RL, which optimizes expected reward and has been shown to improve single-shot (pass@1) accuracy~\citep{shao2024deepseekmath,yue2025does}.

Our RL implementation adopts a multi-task setup (illustrated in Appendix Figure~\ref{fig:rl}):
50\% of the inputs are used for whole proof generation, and the remaining 50\% for first-round self-correction. We train for a single epoch, using approximately 46K and 67K unique inputs for the 8B and 32B models, respectively. 
On the algorithmic side, we use a hybrid GRPO-based approach~\citep{shao2024deepseekmath}. 
Compared to vanilla GRPO, our method removes group normalization as suggested by Dr.GRPO~\citep{liu2025understanding} to avoid inherent bias on length, incorporates clip-higher, overlong penalties, and dynamic sampling from DAPO~\citep{yu2025dapo}, and excludes the KL regularization term from the objective to encourage exploration.
A key observation is that question difficulty significantly impacts RL training. To address this, we modify the dynamic sampling strategy to only include problems with pass rates in the range (0, 0.75] during optimization. For further implementation details, see Appendix~\ref{app:rl_details}.





\paragraph{Model averaging for enhanced diversity.}
We observed that in the later stages of SFT and RL, the model's diversity decreases. This is reflected by an increase in pass@1 but a decline in pass@N for larger values of N, such as N=32. We adopts model averaging to enhance model diversity~\citep{wortsman2022robust, wortsman2022model, lin2023mitigating, lin2023spurious, dang2025assessing}
. Specifically, let the parameters of the base model be denoted as $\theta_0$, and those of the fine-tuned model as $\theta$. We use the combined model parameters defined as $(1-\alpha)\theta_0 + \alpha \theta$, where $\alpha \in (0, 1)$. Existing literature has demonstrated that this simple approach can significantly improve the feature diversity of the final model. Our observations also confirm that this method effectively enhances pass@N. We perform model averaging at each stage of the process. Specifically, we apply model averaging after completing SFT and use the averaged model as the starting point for RL. Once RL is completed, we perform model averaging again and use this averaged model as the final model.

\subsection{Whole pipeline: putting everything together}
\label{sect:overall_pipeline}
\begin{figure}[t]
    \centering
    \includegraphics[width=1.0\linewidth]{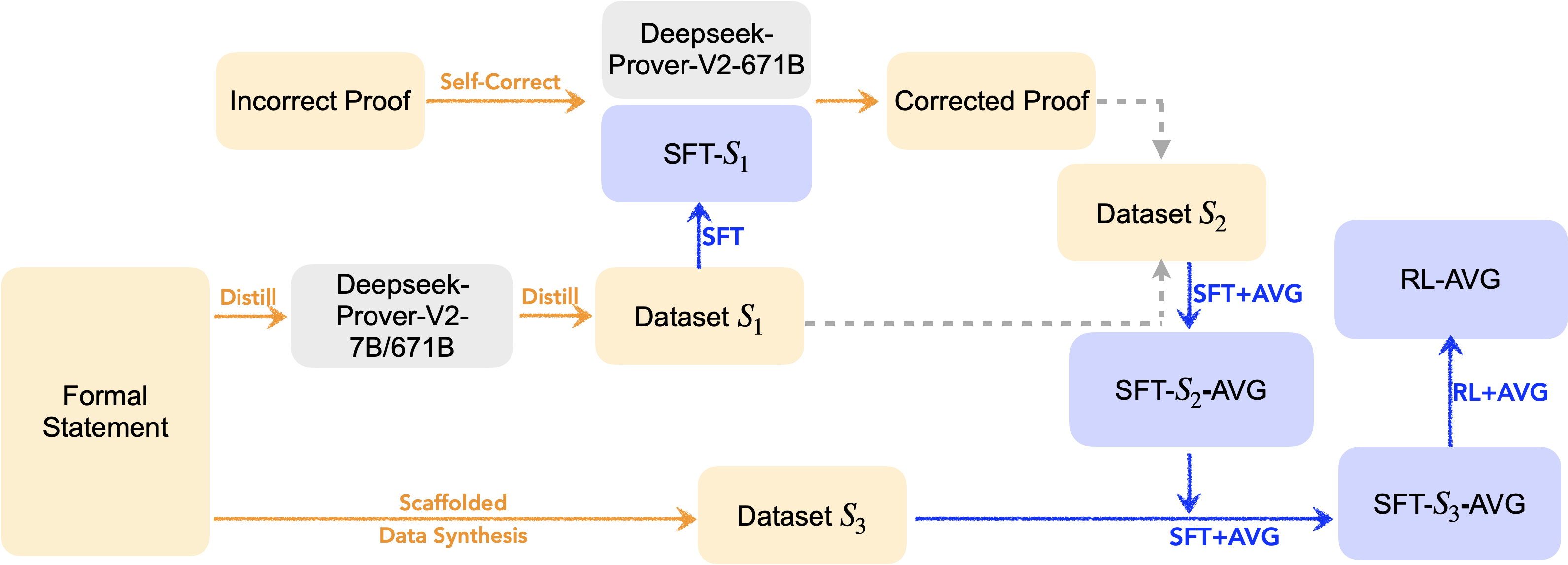}
    \caption{The overall workflow of model training. ``+AVG" denotes that the trained model is averaged with the base model after training. ``RL-AVG" is the final output model. Detailed descriptions are provided in Section~\ref{sect:overall_pipeline}.}
    \label{fig:work_flow}
\end{figure}

The pipeline consists of the following steps:

\begin{itemize}
    \item 
    We utilize {Deepseek-Prover-V2-7B} and {Deepseek-Prover-V2-671B} to perform large-scale inference, producing an initial supervised fine-tuning (SFT) dataset $S_1$ for complete proof generation.

    \item 
    We employ $S_1$ to conduct SFT on both {Qwen3-8B} and {Qwen3-32B}, resulting in fine-tuned models SFT-$S_1$ for both the 8B and 32B variants\footnote{We use \url{https://github.com/hiyouga/LLaMA-Factory} for SFT.}.

    \item 
    Using SFT-$S_1$ and {Deepseek-Prover-V2-671B}, we annotate self-correction data. The self-correction data from Deepseek-Prover-V2-671B is inferenced using NeMo-Skills\footnote{\url{https://github.com/NVIDIA/NeMo-Skills}} with 144 H100s. This data is then incorporated back into $S_1$ to create an enhanced dataset $S_2$. We subsequently perform SFT on both model sizes using $S_2$, yielding improved models SFT-$S_2$. We perform model averaging between SFT-$S_2$ and the base model for both the 8B and 32B models, resulting in SFT-$S_2$-AVG.

    \item 
    We perform scaffolded data synthesis with SFT-$S_2$-AVG to generate the dataset $S_3$. Continuing the SFT process, we further improve the models by training SFT-$S_2$-AVG on $S_3$, yielding SFT-$S_3$, and subsequently obtain the averaged model SFT-$S_3$-AVG.

    \item 
    We apply reinforcement learning to the SFT-$S_3$-AVG and conduct model averaging and obtain the final model RL-AVG.
\end{itemize}

The overall workflow is illustrated in Fig.~\ref{fig:work_flow}.

%% file: sections/experiments.tex
\section{Evaluation}\label{sec:eval}

This section presents our experiment results on \provernew. We first discuss our evaluation benchmarks (\Cref{sec:benchmark}). Then, we discuss our main evaluation results on the selected benchmarks (\Cref{sec:results}), and correspondingly the scaling behavior (\Cref{sec:scaling}). Finally, we investigate the performance of reinforcement learning and model averaging (\Cref{sec:rlavg}).

\subsection{Benchmarks}\label{sec:benchmark}


\paragraph{\miniff.} \miniff~\citep{zheng2021minif2f} consists of 488
problem statements (244 validation and 244 test problems) in Lean. The problems are
drawn from high-school level competitions including
the AMC, AIME, and the International Mathematical Olympiad (IMO). We use the version of MiniF2F provided by Kimina~\citep{wang2025kimina}\footnote{\url{https://huggingface.co/datasets/AI-MO/minif2f_test}}, which has some incorrect statements fixed.

\paragraph{\putnam.} \putnam~\citep{tsoukalas2024putnambench} focuses on college-level mathematics competition problems that are sourced from the William Lowell Putnam Mathematical
Competition years 1962 - 2023. \putnam comprises 644 problems, covering
algebra, analysis, number theory, geometry, combinatorics, probability, and set theory.

\paragraph{\mobench.}

\begin{figure}[!t]
    \centering
    \begin{minipage}[b]{0.32\textwidth}
        \centering
        \includegraphics[width=\textwidth]{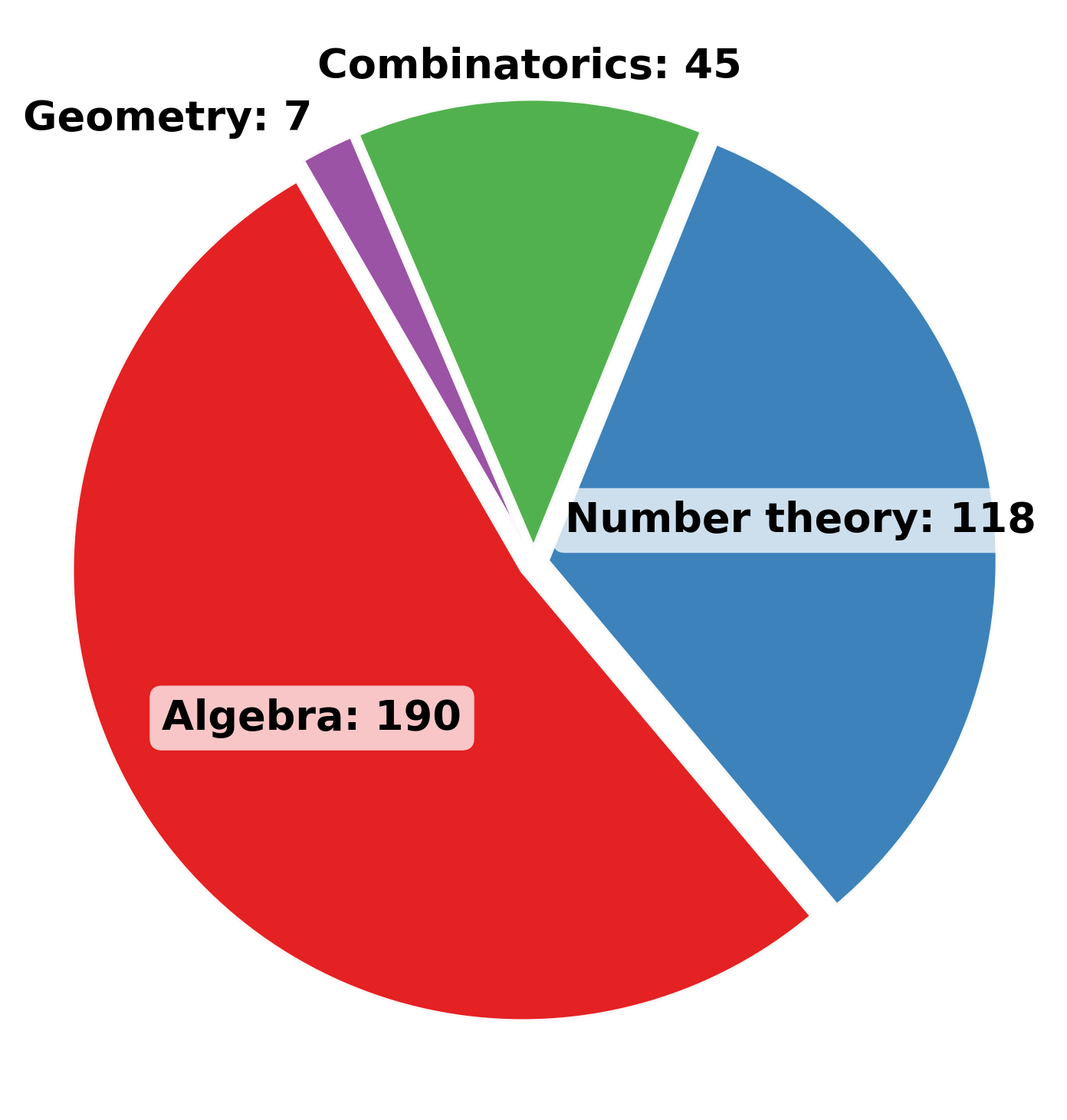}
        \caption{Distribution of problems in \mobench by category.}
        \label{fig:mobench:pie}
    \end{minipage}
    \hspace{0.02\textwidth}
    \begin{minipage}[b]{0.64\textwidth}
        \centering
        \includegraphics[width=\textwidth]{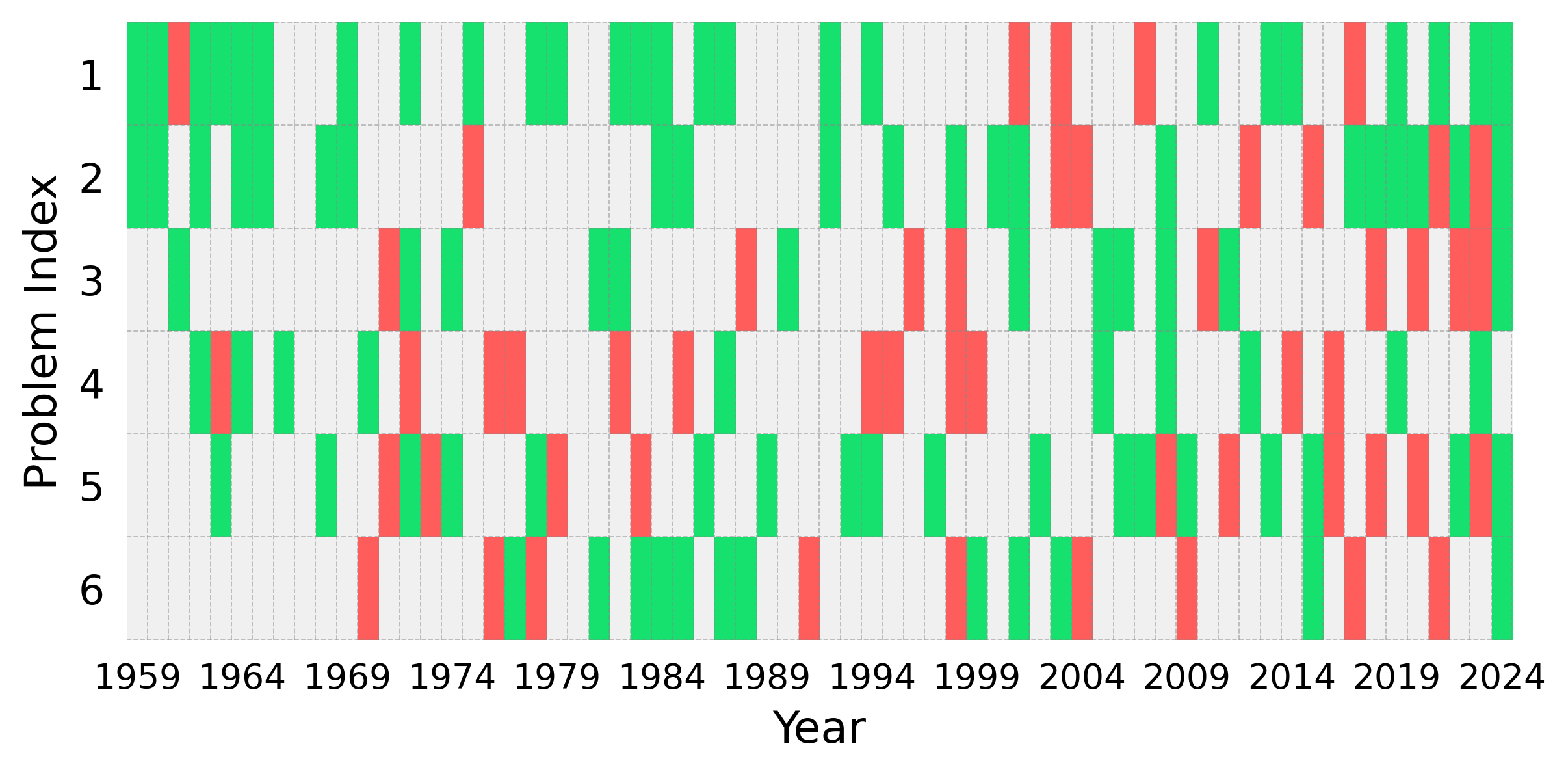}
        \caption{
            Community's achievement on IMO problems by year and problem index. Each cell represents a problem: 
            \textcolor{gray}{\rule{10pt}{8pt}} statement not formalized, 
            \textcolor{red}{\rule{10pt}{8pt}} not solved, and 
            \textcolor{green!60!black}{\rule{10pt}{8pt}} solved.
        }
        \label{fig:mobench:human}
    \end{minipage}
\end{figure}

We constructed \mobench, which comprises 360 human-verified formalizations of Olympiad-level mathematical problems, sourced from Compfiles~\footnote{\url{https://dwrensha.github.io/compfiles/imo.html}} and IMOSLLean4 repository~\footnote{\url{https://github.com/mortarsanjaya/IMOSLLean4}}. It contains 158 IMO problems from 1959 to 2024, 131 IMO shortlist problems covering 2006 to 2023, 68 national mathematical Olympiad problems, and 3 additional mathematical puzzles. The statistic of problem categories in \mobench is presented in Figure~\ref{fig:mobench:pie}, and Figure~\ref{fig:mobench:human} visualizes humans' formalizing and solving status of IMO problems in \mobench. See Appendix~\ref{app:case_studies} for more details of \mobench and its comparison with \miniff.

\subsection{Evaluated Methods}

Following previous works~\citep{xin2024deepseekv15,lin2025goedel,dong2025stp,ren2025deepseek} the evaluations are done under Lean 4.9.0-rc1. For the first round of whole-proof generation, the max token length of the model is set to be 30,000. For the verifier-guided error-correction, we sequentially conduct 2 additional rounds of self-correction, given the verifier's feedback on the previous attempt. The total number of tokens in the self-correction mode is set to be 40,000. We report the pass@N metric.

\subsection{Main results}\label{sec:results} 

\input{tables/minif2f}

The evaluation results of Goedel-Prover-V2 on \miniff are summarized in \Cref{tab:minif2f}, and results on \putnam are shown in \Cref{tab:putnam}. The results for \mobench are presented in the rightmost figure of \Cref{fig:combined-performance}. Below, we summarize and discuss the results.

\paragraph{High performance at modest scale.}
Our 32B model achieves pass@32 accuracy of 88.1\% on \miniff, with 90.4\% after 2 rounds of error-correction, exceeding the previous state-of-the-art DeepSeek‑Prover‑V2‑671B’s 82.4\% while using far fewer parameters. Even the 8B variant achieves 84.6\%, nearly matching or outperforming Kimina‑Prover‑70B’s results under the same inference budget, and outperforming the previous SOTA DeepSeek-Prover-V2-671B on \miniff, with an significant smaller model size. On \putnam, our 32B model solves 43 problems under pass@32, nearly doubling the DeepSeek-Prover-V2-671B model's performance under the same budget. With error-correction, Goedel-Prover-V2-32B solves 57 problems under pass@32, outperforming DeepSeek-Prover-V2-671B under pass@1024. Under pass@184 and error correction, Goedel-Prover-V2-32B solves 86 on \putnam, securing the best open-source theorem prover on the leaderboard.

\input{tables/putnam}

\paragraph{Efficacy of verifier‑guided self‑correction.}
Adding self‑correction provides a consistent gain of approximately 2 percentage points in pass@32 for both models on \miniff. On \putnam, error correction leads to 14 more solves under pass@32. This demonstrates that integrating Lean compiler feedback into a long-chain-of-thought revision pipeline enables the model to identify errors and repair them effectively.

\paragraph{Sample-efficient inference.}
Unlike prior models such as Kimina‑Prover and DeepSeek‑Prover-V2, which rely heavily on large sampling budgets or test-time reinforcement learning to reach peak accuracy, Goedel‑Prover‑V2 attains very high pass@N with minimal inference overhead (N=32 or 64), indicating that the model internalizes powerful reasoning strategies during training. The sample efficiency, together with the relatively small size, makes Goedel-Prover-V2 series a very good candidate for the community to develop new algorithms and test on different benchmarks.

\subsection{Scaling Analysis}\label{sec:scaling}


\begin{wrapfigure}{r}{0.4\textwidth}
\vspace{-15mm}
    \centering
    \includegraphics[width=\linewidth]{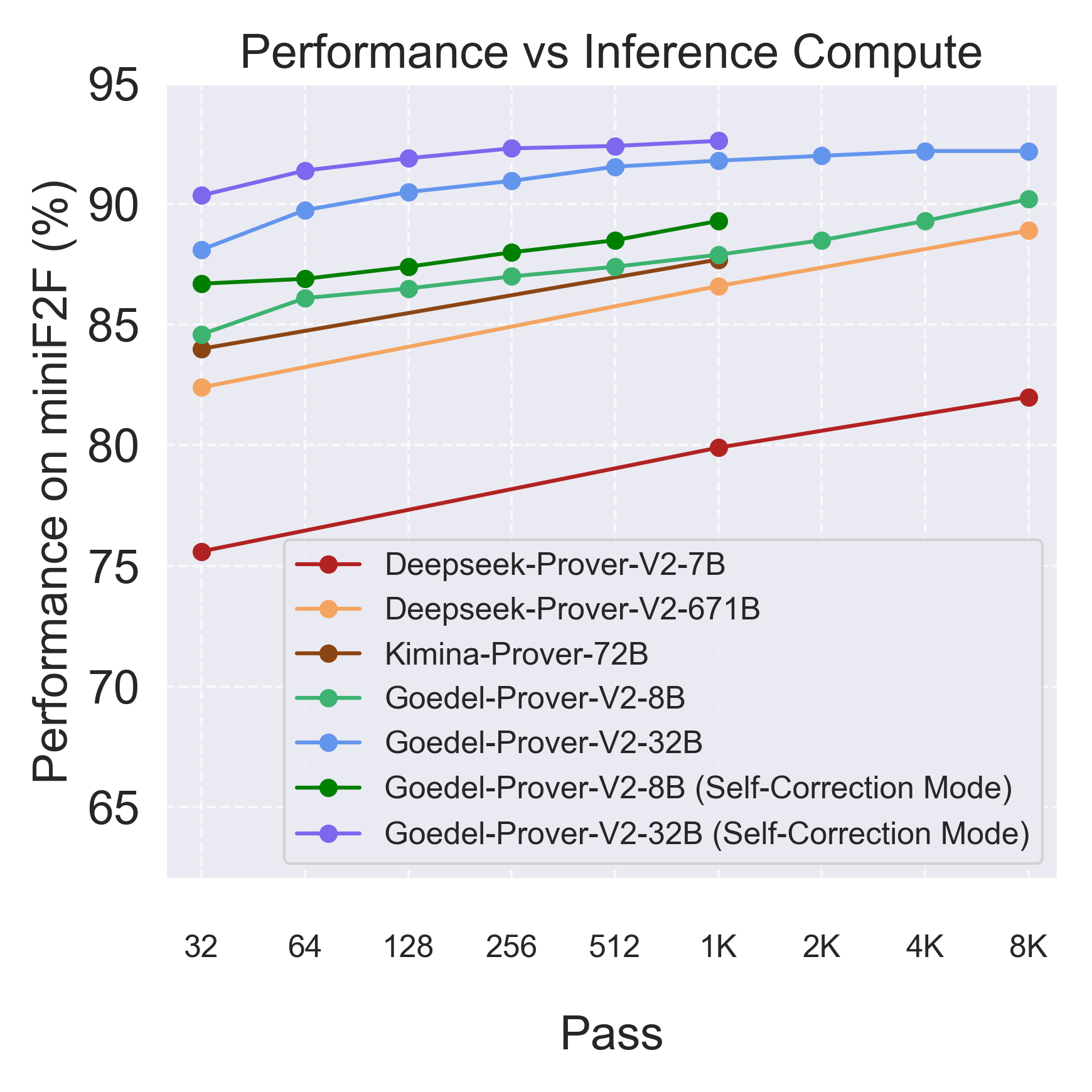}
    \vspace{-5mm}
    \caption{Scaling behavior on \miniff test split. }
    \label{fig:scaling}
\vspace{-15mm}
\end{wrapfigure}

\Cref{fig:scaling} and \Cref{tab:performance-horizontal} illustrate the scaling behavior of Goedel-Prover-V2 (8B and 32B variants) across different inference budgets, compared against DeepSeek-Prover-V2 (7B and 671B) and Kimina-Prover-72B. At the lower sampling regime (pass@32), Goedel-Prover-V2-32B already achieves 88.1\%, notably surpassing DeepSeek-Prover-V2-671B (82.4\%) and Kimina-Prover-72B (84.0\%). This advantage persists across inference budgets, with our self-correction mode providing approximately a 2-point performance improvement under pass@32 and pass@64, peaking at 92.6\% at pass@8192. The smaller 8B model also demonstrates strong scalability, surpassing the 671B DeepSeek model at all budgets.

\begin{table}[!ht]
\centering
\small
\begin{tabular}{lccccccccc}
\toprule
\textbf{Model} & \textbf{32} & \textbf{64} & \textbf{128} & \textbf{256} & \textbf{512} & \textbf{1024} & \textbf{2048} & \textbf{4096} & \textbf{8192} \\
\midrule
32B (self-correction mode) & 90.4 & 91.4 & 91.9 & 92.3 & 92.4 & 92.6 & -- & -- & -- \\
32B            & 88.1 & 89.8 & 90.5 & 91.0 & 91.6 & 91.8 & 92.0 & 92.2 & 92.2 \\
8B (self-correction mode) & 86.7 & 86.9 & 87.4 & 88.0 & 88.5 & 89.3 & -- & -- & -- \\
8B            & 84.6 & 86.1 & 86.5 & 87.0 & 87.4 & 87.9 & 88.5 & 89.3 & 90.2 \\
\bottomrule
\end{tabular}
\caption{The performance (\%) of Goedel-Prover-V2 on \miniff across different compute budget.}
\label{tab:performance-horizontal}
\end{table}


These results indicate that Goedel-Prover-V2 efficiently internalizes reasoning during training, requiring fewer inference samples to achieve comparable or superior accuracy. Furthermore, the consistent gains from verifier-guided self-correction across all inference budgets further underscore the value of combining error correction with long-chain-of-thought reasoning in formal theorem proving.

\begin{wrapfigure}{r}{0.5\textwidth}
\vspace{-5mm}
    \centering
    \includegraphics[width=\linewidth]{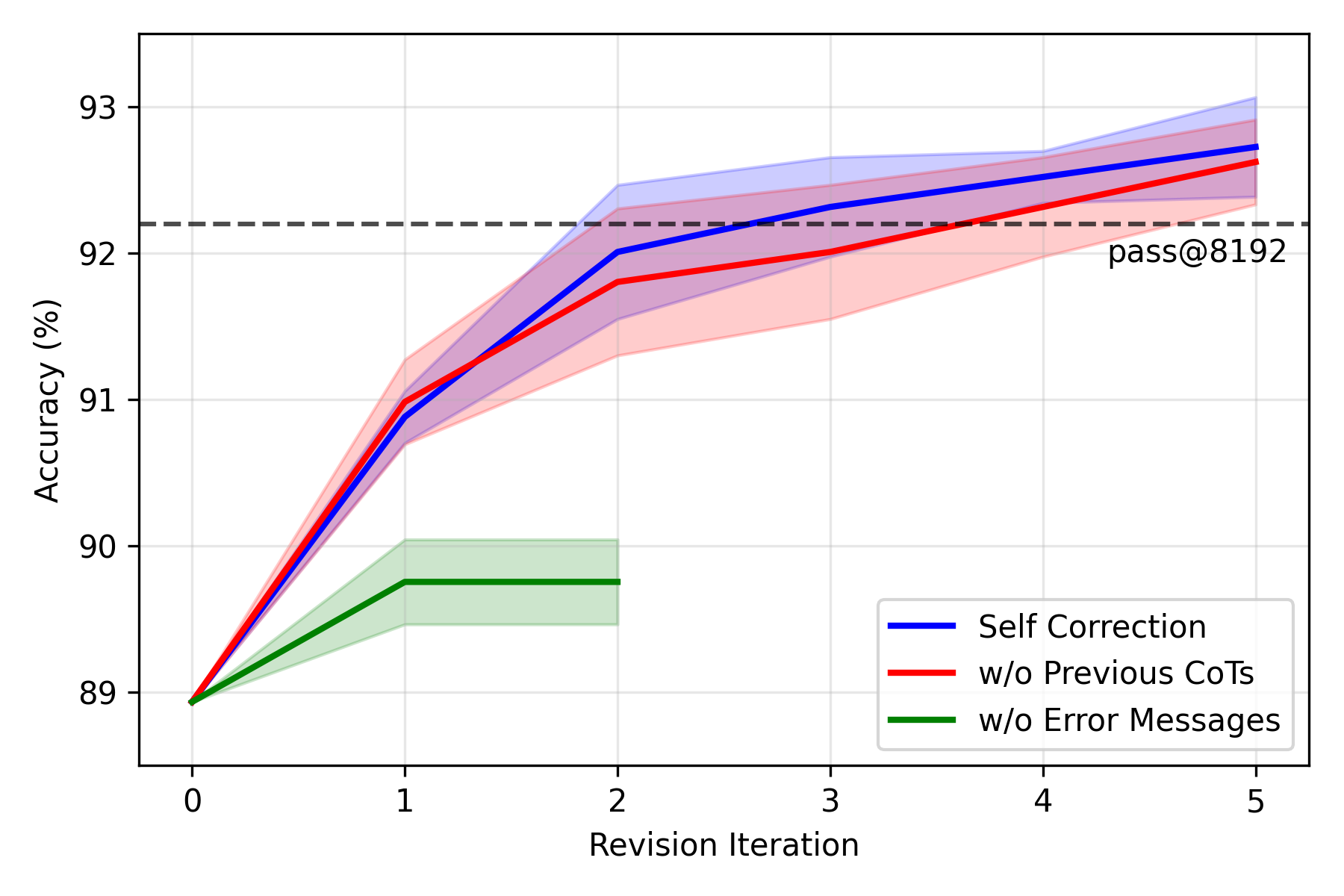}
    \vspace{-8mm}
\caption{Ablation study on self-correction with extended context length and revision iterations on the \miniff test split at pass@32.}
\label{fig:self-correction-ablation}
\vspace{-5mm}
\end{wrapfigure}

\subsection{Analysis of Self-Correction}
In our main experiments, self-correction was conducted for a maximum of 2 rounds with a 40k token context length. To further explore its capabilities, we used YaRN~\citep{peng2024yarn} to extend the context length to 128k tokens and allowed up to 5 revision iterations. We conducted a series of experiments on the \miniff benchmark at pass@32, the results of which are presented in \Cref{fig:self-correction-ablation}. Alongside our standard prompting setup (Self Correction), we performed two ablation studies: (1) removing the specific compiler error messages (w/o Error Messages), and (2) removing the chain-of-thought from previous attempts, retaining only the formal proof (w/o Previous CoTs).

The results show that removing compiler feedback significantly lowers performance, confirming that specific error messages are crucial for effective revision. Similarly, removing the reasoning from previous attempts also slightly degrades performance, indicating that retaining the chain-of-thought from prior rounds is beneficial. Furthermore, with an extended context and more revision iterations, the full self-correction model's pass@32 accuracy on \miniff reaches an average of 92.7\%, which surpasses the 92.2\% performance of the model without self-correction at pass@8192, highlighting the sample efficiency of our iterative revision process.

\subsection{RL and Model averaging}\label{sec:rlavg}

\begin{figure}
    \centering
    \includegraphics[width=\linewidth]{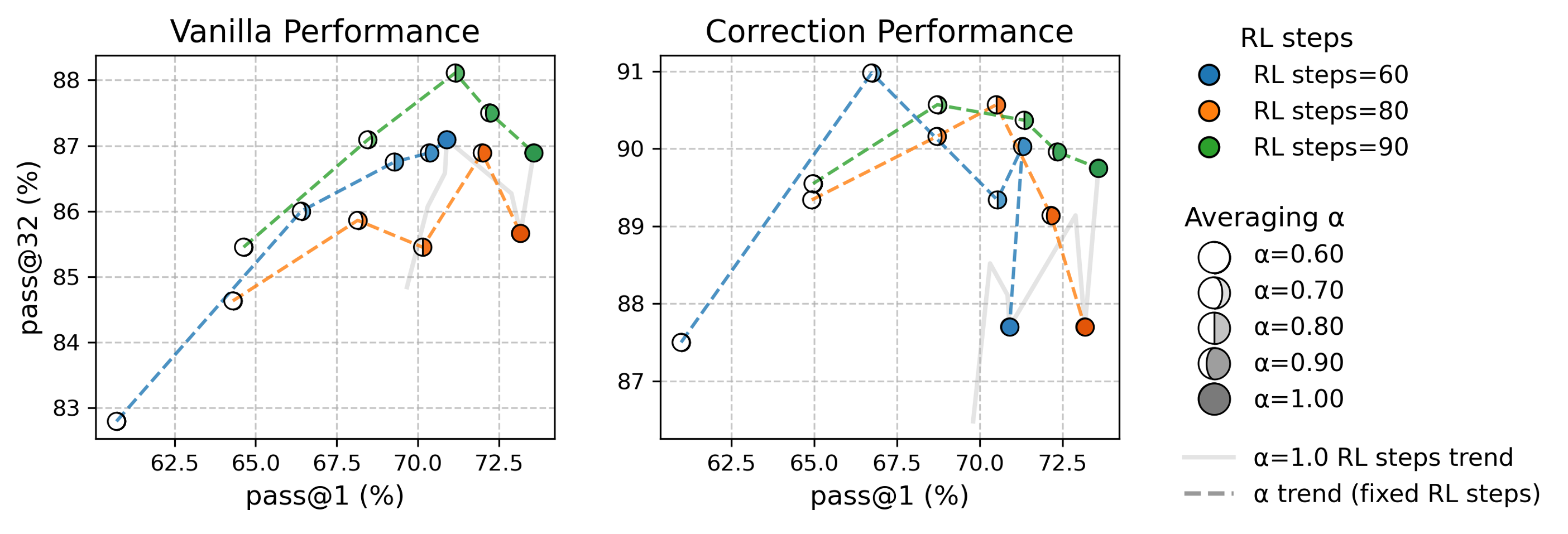}
    \caption{The effects of varying RL steps and model averaging ratios on the pass@1 and pass@32 performance of models, both with and without correction.}
    \label{fig:rl_avg}
\end{figure}

We systematically evaluate the impact of RL steps and model averaging strategies on the performance of Goedel-Prover-V2. Specifically, for RL checkpoints at steps 60, 80, and 90, we apply model averaging with coefficients $\alpha = 0.6, 0.7, 0.8, 0.9$ (where $\alpha$ is the weight of the base model). We assess each averaged model in both vanilla (whole-proof generation) and correction (with self-correction) settings, evaluating both $\text{pass@1}$ and $\text{pass@N}$, as visualized in Figure~\ref{fig:rl_avg}.

For both vanilla and correction settings, $\text{pass@1}$ consistently increases as the number of RL steps grows. For vanilla $\text{pass@N}$, performance stabilizes at higher RL steps, whereas in the correction setting, $\text{pass@N}$ continues to improve. This indicates that correction benefits more from RL, likely due to the shortage of high-quality self-correction data in the SFT stages.

Examining different values of $\alpha$, we observe that a higher proportion of the base model (i.e., lower $\alpha$) leads to lower $\text{pass@1}$, but $\text{pass@N}$ first rises and then falls as $\alpha$ increases. There exists an optimal model averaging ratio that maximizes $\text{pass@N}$. This trend holds for both vanilla and correction settings, with the improvement being more pronounced for correction. This suggests that model averaging not only improves sample diversity but also amplifies the benefits of RL-driven self-correction.

%% file: tables/minif2f.tex
\begin{table}[!th]
\small
    \centering
    \caption{Performance of different whole-proof generation methods on \miniff test split. \textsuperscript{\dag} denotes concurrent work}
    \label{tab:minif2f}
    \begin{tabular}{lcc}
    \toprule
        Method & Budget & Performance \\
    \midrule
        \prover~\citep{lin2025goedel} & 32 & 57.6\% $\pm$ 0.7\%\\
          & 3200 & 62.7\% \\
    \midrule
        STP~\citep{dong2025stp} & 128 & 61.2\% $\pm$ 0.6\%\\
          & 3200 & 65.0\% $\pm$ 0.5\%\\
          & 25600 & 67.6\% \\
    \midrule
        Kimina-Prover-Preview-72B~\citep{wang2025kimina} & 32 & 68.85\% \\
          & 8192 & 80.74\% \\
    \midrule
        DeepSeek-Prover-V2-7B~\citep{ren2025deepseek} & 32 & 75.6\% $\pm$ 0.5\% \\
          & 8192 & 82.0\% \\
          \hdashline[4pt/2pt]
        DeepSeek-Prover-V2-671B & 32 & 82.4\% $\pm$ 0.6\% \\
          & 8192 & 88.9\% \\
    \midrule
        Kimina-Prover-8B\textsuperscript{\dag}~\citep{wang2025kimina} & 32 & 78.3\% \\
        \hdashline[4pt/2pt]
        Kimina-Prover-70B\textsuperscript{\dag} & 32 & 84.0\% \\
          & 1024 & 87.7\% \\
        \hdashline[4pt/2pt]
        \quad\quad w/ TTRL & unknown & 92.2\% \\
    \midrule
        \rowcolor{gray!15}\provernew-8B & 32 & 84.6\% $\pm$ 0.6\% \\
        \rowcolor{gray!15} & 1024 & 87.9\%  \\
        \rowcolor{gray!15} & 8192 & 90.2\%  \\
        \hdashline[4pt/2pt]
        \rowcolor{gray!15}\quad\quad w/ self-correction & 32 & 86.7\% $\pm$ 0.2\% \\
        \rowcolor{gray!15}\quad\quad  & 1024 & 89.3\%   \\
        \hdashline[4pt/2pt]
        \rowcolor{gray!15}\provernew-32B & 32 & 88.1\% $\pm$ 0.8\% \\
        \rowcolor{gray!15} & 1024 & 91.8\%  \\
        \rowcolor{gray!15} & 8192 & 92.2\%  \\
        \hdashline[4pt/2pt]
        \rowcolor{gray!15}\quad\quad w/ self-correction & 32 & 90.4\% $\pm$ 0.6\%  \\
        \rowcolor{gray!15}\quad\quad  & 1024 & 92.6\%   \\
    \midrule
    \bottomrule
    \end{tabular}
\end{table}

%% file: tables/putnam.tex

\begin{table}[!th]
\small
\centering
\begin{threeparttable}
\caption{Comparison of different models on PutnamBench. Our Goedel-Prover-V2 with compiler-guided self-correction solves 86 problems from PutnamBench, improving the previous SOTA (DeepSeek-Prover-V2) by 39 more problems, and securing the best open-source model on the leaderboard.\tnote{*}}
\label{tab:putnam}
\begin{tabular}{llcccl}
\toprule
\textbf{\#} & \textbf{Model} & \textbf{num-solved} & \textbf{open-source} & \textbf{compute} \\
\midrule
1 & \textbf{Goedel-Prover-V2 (self-correction mode)} & 86 & \checkmark & pass@184 \\
1 & \textbf{Goedel-Prover-V2 (self-correction mode)} & 57 & \checkmark & pass@32 \\
1 & \textbf{Goedel-Prover-V2} & 43 & \checkmark & pass@32 \\
2 & DeepSeek-Prover-V2 & 47 & \checkmark & pass@1024 \\
2 & DeepSeek-Prover-V2 & 22 & \checkmark & pass@32 \\
3 & DSP+ & 23 & \checkmark & pass@128 \\
4 & Bourbaki & 14 & \checkmark & pass@512 \\
5 & Kimina-Prover-7B-Distill & 10 & \checkmark & pass@192 \\
6 & Self-play Theorem Prover & 8 & \checkmark & pass@3200 \\
7 & Goedel-Prover-SFT & 7 & \checkmark & pass@512 \\
8 & ABEL & 7 & $\times$ & pass@596 \\
\bottomrule
\end{tabular}
\begin{tablenotes}
    \item[*] A concurrent work, Seed Prover~\citep{chen2025seedproverdeepbroadreasoning}, successfully solved 331 problems on PutnamBench. However, the prover is not open-source, and it is not clear what the computational budget is at test time (which is expected to be much larger than ours).
\end{tablenotes}
\end{threeparttable}
\end{table}

%% file: sections/related.tex
\section{Related Works}\label{sec:related}

\paragraph{Formal Theorem Proving as Whole Proof Generation.} 
Inspired by the success of end-to-end reasoning in informal LLMs, recent foundational work in formal theorem proving has adopted a similar strategy: generating entire Lean proofs in a single pass. While informal reasoning models often struggle with step verification—producing fluent yet logically flawed arguments~\citep{petrov2025proof}—formal settings like Lean~\citep{de2015lean} offer a precise and executable verifier that enforces correctness. Leveraging this, models such as DeepSeek-Prover~\citep{xin2024deepseek}, Goedel Prover~\citep{lin2025goedel}, and Kimina Prover~\citep{wang2025kimina} demonstrate that end-to-end generation can produce formally verified proofs that are immediately checkable by the Lean proof assistant. This paradigm retains the global coherence and simplicity of sequence generation while grounding outputs in verifiable formal logic.

\paragraph{Formal Theorem Proving using Proof Search.}
In contrast to end-to-end generation, proof search–based methods guide the model to incrementally construct proofs by exploring derivation paths with verifier feedback at each step. Recent systems such as InternLM2.5-StepProver~\citep{wu2024internlm2}, Hunyuan Prover~\citep{li2024hunyuanprover}, DeepSeek-Prover v1.5~\citep{xin2024deepseekv15}, and BFS Prover~\citep{xin2025bfs} employ tree search algorithms---such as Monte Carlo Tree Search or breadth-first search---to explore multiple proof trajectories and iteratively assemble valid proofs. While this strategy improves correctness, it comes at the cost of significantly higher computational overhead, as the model must evaluate and verify a large number of partial branches. Recently, hybrid methods, which use a general-purpose (strong) LLM to write the proof sketch and query theorem provers to fill in the proof for small steps, also emerged (DSP+~\citep{cao2025reviving} and the concurrent work Delta-Prover~\citep{zhou2025solving}). However, these works usually require the LLM to be very large and powerful, where DSP+ uses 671B DeepSeek-R1~\citep{guo2025deepseek} or DeepSeek-V3~\citep{liu2024deepseek}, and Delta-Prover uses Gemini~\citep{comanici2025gemini}. Notably, Seed-Prover~\citep{chen2025seedproverdeepbroadreasoning} employs a wide range of techniques, including extensive test-time search and refinement, and achieves IMO medal-level performance. However, it requires substantial computational resources.



\paragraph{Self-Repair and Verifier-Guided Refinement.} While tree-based proof search improves correctness, its high computational cost has motivated the development of more efficient mechanisms for guiding search and enabling self-refinement. One direction explores how auxiliary signals or intermediate structures can streamline the reasoning process. \citet{jiang2023draft, gloeckle2024abel, cao2025reviving} bridge informal and formal reasoning by using informal sketches as a skeleton for formal proof generation, while \citet{yang2024leandojo} employs retrieval-augmented generation (RAG) to retrieve relevant theorems from formal libraries during proof construction. Other approaches, such as \citet{ji2025leanabell}, implement self-verification and iterative refinement loops, allowing models to autonomously revise candidate proofs using verifier feedback.

Verifier-in-the-loop strategies in recent works~\citep{baba2025prover, zhou2025solving, ren2025deepseek, wang2025kimina,chen2025seedproverdeepbroadreasoning}, also highlight the benefits of interactive feedback, improving success rates through iterative correction rather than brute-force search. These methods collectively point to a more flexible and scalable paradigm: enabling the model to diagnose and repair its own outputs with minimal external supervision.

Goedel-Prover-V2 builds on this trajectory by adopting a self-revision framework, where the model iteratively proposes and refines candidate proofs until they satisfy the Lean checker. This architecture draws inspiration from general-purpose self-repair frameworks in coding and reasoning~\citep{yao2023tree, shinn2023reflexion, first2023baldur, olausson2024is, chen2024teaching, bouzenia2024repairagent}, bringing their iterative refinement loop into the domain of formal mathematics with long chain-of-thought reasoning.





%% file: sections/conclusion.tex
\section{Conclusion and Discussion}

In this work, we introduced \textbf{Goedel-Prover-V2}, a series of state-of-the-art open-source theorem provers that significantly advance automated formal proof generation. Our key innovation is the integration of long-chain-of-thought reasoning with compiler-guided self-correction, addressing a crucial gap overlooked by prior methods. Through scaffolded data synthesis, SFT and RL training, and model averaging, our models achieve state-of-the-art performance among open-source provers. In particular, our 32B model achieves 88.1\% on MiniF2F at pass@32 and benefits further from self-correction and reaching 90.4\% on MiniF2F at pass@32, outperforming larger models with lower inference cost. Moreover, our 8B model also outperforms the previous SOTA DeepSeek-Prover-V2-671B under pass@32 on MiniF2F with a significant smaller model size.


We also investigate a proof repair strategy as a test-time alternative to standard pass@k sampling. Instead of regenerating an entire failed proof, our method corrects only the faulty segment. We use Lean 4’s compiler feedback and the $\texttt{extract\_goal}$ tactic to isolate the unsolved subgoal, prompt the prover to solve it independently, and then reinsert the correct solution into the original proof. On the MiniF2F benchmark, this approach improves the amortized budget scaling curve by 1–2 percentage points, highlighting inference-time scaling strategies as a key direction for future work.

By open-sourcing all trained models, we aim to catalyze further advancements in formal reasoning research. We envision that the release of Goedel-Prover-V2 will not only establish a new benchmark for automated theorem proving but also provide a practical, efficient platform upon which future innovations can be built.

\section*{Acknowledgement}

This project was the result of a close collaborative effort by all authors, and every major component benefited from joint discussion and iteration. The author would like to thank Igor Gitman for support with the curation of self-correction data.

HZ and SA acknowledge the support from Schmidt, Darpa, ONR, and NSF. CJ acknowledges the support from NSF-OAC-2411299, NSF-IIS-2239297, and Princeton AI Lab Seed Fund.

%% file: sections/appendix.tex
\section*{Appendix}

\section{Details of \mobench}
\label{app:case_studies}

\mobench is human-processed to eliminate several issues presented in the source problems: 1. incomplete problem statements, 2. distribution across multiple files, 3. multiple theorems per problem, and 4. incompatibility with the commonly used Mathlib. The verification process ensures that each problem contains exactly one formal theorem with its corresponding informal statement, and confirms that all formal statements can pass the compilation with the \texttt{sorry} tactic.

We compared the IMO problems shared between \mobench and \miniff, and identified at least 3 cases in \miniff exhibiting issues such as: 1. the formal statement to be proved is strictly weaker than the informal statement, and 2. the formal statement does not match the informal statement. Notably, similar issues are not observed for these problems in \mobench.

In the following, we present three case studies comparing the problem formalizations in \miniff and \mobench.

\subsection{IMO 1981, Problem 6: Specific Value vs. General Formula}

\vspace{1em}

\begin{minipage}[t]{0.48\textwidth}
    \begin{leancode}[title=\miniff]
/--
The function $f(x,y)$ satisfies

(1) $f(0,y)=y+1, $

(2) $f(x+1,0)=f(x,1), $

(3) $f(x+1,y+1)=f(x,f(x+1,y)), $

for all non-negative integers $x,y $. Determine $f(4,1981) $.
-/

theorem imo_1981_p6 (f : ℕ → ℕ → ℕ) 
  (h₀ : ∀ y, f 0 y = y + 1) 
  (h₁ : ∀ x, f (x + 1) 0 = f x 1)
  (h₂ : ∀ x y, f (x + 1) (y + 1) = 
        f x (f (x + 1) y)) : 
  ∀ y, f 4 (y + 1) = 2 ^ (f 4 y + 3) - 3 := by
\end{leancode}
\end{minipage}
\hfill
\begin{minipage}[t]{0.48\textwidth}
    \begin{leancode}[title=\mobench]
/-!
# International Mathematical Olympiad 1981, Problem 6

Suppose that f : ℕ × ℕ → ℕ satisfies

 1) f (0, y) = y + 1
 2) f (x + 1, 0) = f (x, 1),
 3) f (x + 1, y + 1) = f (x, f (x + 1, y))

for all x y ∈ ℕ.

Determine f (4, 1981).
-/

def no_eval (x : ℕ) : ℕ := x
abbrev solution_value : ℕ := 
  no_eval ((2^·)^[1984] 1 - 3)

theorem imo1981_p6 (f : ℕ → ℕ → ℕ)
  (h1 : ∀ y, f 0 y = y + 1)
  (h2 : ∀ x, f (x + 1) 0 = f x 1)
  (h3 : ∀ x y, f (x + 1) (y + 1) = 
        f x (f (x + 1) y)) :
  f 4 1981 = solution_value := sorry
    \end{leancode}
\end{minipage}

The informal statement requires computing the exact value of \( f(4, 1981) \), a specific number. However, the \miniff formalization only asks to prove a general recurrence relation, which is an intermediate step in the solution process. This discrepancy can make the formal proof substantially easier than solving the original problem. In contrast, \mobench faithfully translates the original problem into formal language, requiring the proof of the final numerical value as demanded by the IMO statement.

\subsection{IMO 1983, Problem 6: Incomplete vs. Full Condition}

\vspace{1em}

\begin{minipage}[t]{0.48\textwidth}
    \begin{leancode}[title=\miniff]
/-- Let $a$, $b$ and $c$ be the lengths of the sides of a triangle. Prove that

$a^2 b(a-b) + b^2 c(b-c) + c^2 a(c-a) \geq 0$.

Determine when equality occurs.
-/

theorem imo_1983_p6 (a b c : $\mathbb{R}$) 
  (h₀ : 0 < a ∧ 0 < b ∧ 0 < c) 
  (h₁ : c < a + b) (h₂ : b < a + c)
  (h₃ : a < b + c) : 
  0 ≤ a^2*b*(a-b) + b^2*c*(b-c) + c^2*a*(c-a) := by
    \end{leancode}
\end{minipage}
\hfill
\begin{minipage}[t]{0.48\textwidth}
    \begin{leancode}[title=\mobench]
/-!
# International Mathematical Olympiad 1983, Problem 6

Suppose that a,b,c are the side lengths of a triangle. Prove that

   $a^{2}b(a - b) + b^{2}c(b - c) + c^{2}a(c - a) \geq 0.$

Determine when equality occurs.
-/

abbrev EqualityCondition (a b c : $\mathbb{R}$) : Prop := 
  a = b ∧ a = c
  
theorem imo1983_p6 
  (T : Affine.Triangle $\mathbb{R}$ (EuclideanSpace $\mathbb{R}$ (Fin 2))) :
  let a := dist (T.points 1) (T.points 2)
  let b := dist (T.points 0) (T.points 2)
  let c := dist (T.points 0) (T.points 1)
  0 ≤ a^2*b*(a-b) + b^2*c*(b-c) + c^2*a*(c-a) ∧
  (0 = a^2*b*(a-b) + b^2*c*(b-c) + c^2*a*(c-a) ↔
  EqualityCondition a b c) := sorry
    \end{leancode}
\end{minipage}

The informal statement has two parts: proving an inequality and determining the condition for equality. The \miniff version only formalizes the inequality, completely omitting the second part of the problem. In contrast, \mobench provides a complete formalization.

\subsection{IMO 1962, Problem 2: Informal \& Formal Statement Mismatch}

\begin{minipage}[t]{0.48\textwidth}
    \begin{leancode}[title=\miniff]
/-- Determine all real numbers $x$ which satisfy the inequality:

$\sqrt{\sqrt{3-x}-\sqrt{x+1}}>\dfrac{1}{2}$

Show that it is $\left[\, -1,\quad 1-\frac{\sqrt{127}}{32}\, \right)$.
-/

theorem imo_1962_p2 (x : $\mathbb{R}$) (h₀ : 0 ≤ 3 - x) (h₁ : 0 ≤ x + 1)
    (h₂ : 1 / 2 < Real.sqrt (3 - x) - Real.sqrt (x + 1)) : -1 ≤ x ∧ x < 1 - Real.sqrt 31 / 8 := sorry
    \end{leancode}
\end{minipage}
\hfill
\begin{minipage}[t]{0.48\textwidth}

\begin{leancode}[title=\mobench]
/-!
# International Mathematical Olympiad 1962, Problem 2

Determine all real numbers x which satisfy

$\sqrt{3 - x} - \sqrt{x + 1} > \frac{1}{2}$.
-/

abbrev SolutionSet : Set $\mathbb{R}$ := Set.Ico (-1) (1 - $\sqrt{31}$ / 8)

theorem imo1962_p2 (x : $\mathbb{R}$) :
  x $\in$ SolutionSet $\leftrightarrow$
  x $\leq$ 3 $\land$ -1 $\leq$ x $\land$ 1/2 < $\sqrt{(3 - x)}$ - $\sqrt{(x + 1)}$ := sorry
\end{leancode}
\end{minipage}

For this problem, the original inequality appears in two different versions\footnote{Please refer to: \url{https://artofproblemsolving.com/wiki/index.php/1962_IMO_Problems/Problem_2}.}: one as \(\sqrt{3-x} - \sqrt{x+1} > \frac{1}{2}\) and another as \(\sqrt{\sqrt{3-x}-\sqrt{x+1}} > \frac{1}{2}\). In \miniff, the informal statement uses the latter (the nested square root version), but the formal statement is based on the former (the simpler difference of square roots), resulting in a mismatch between the informal and formal statements. In contrast, \mobench ensures that both the informal and formal statements consistently correspond to the same version of the problem.

\section{Formal Negation}

We attempt to disprove unsolved statements by proving their logical negation. This is achieved by parsing the Lean 4 statements as follows.
\begin{leancode}
theorem fourIsPrime (a : $\mathbb{N}$) (ha : a = 4) : a.Prime := by sorry
theorem fourIsPrimeNeg : ¬ ∀ (a : $\mathbb{N}$) (ha : a = 4), a.Prime := by sorry
\end{leancode}

\section{Details for judging formalization}
\label{app:prompt}

Here is the exact prompt for LLM judging the faithfulness of formalization.

\begin{plainleancode}
You will receive a math problem consisting of its natural language statement along with its formal statement in LEAN 4.

Please evaluate whether the formal LEAN statement appropriately translates the natural language statement based on the following criteria:

1. Key Elements: The problem's essential components are correctly represented in LEAN code.
2. Mathematical Accuracy: The translation preserves the accuracy of the mathematical content.
3. Structural Fidelity: The translation aligns closely with the original problem, maintaining its structure and purpose.
4. Comprehensiveness: All assumptions, conditions, and goals present in the natural language statement are included in the LEAN translation.

Your answer should be in the following format:

Thought: [Your Answer]

Judgement: [Your Answer, one of {Appropriate, Inappropriate}]

---

Following are the example problems label for the reasonability of their translation.

# Example 1:

## Original natural language statement of the problem:

For the graph of a certain quadratic $y = ax^2 + bx + c$, the vertex of the parabola is $(2,10)$, and one of the $x$-intercepts is $(1,0)$.  What is the $x$-coordinate of the other $x$-intercept?

## Translated formal statement:
```lean
theorem quadratic_other_intercept
    (f : $\mathbb{R}$ → $\mathbb{R}$) 
    (a b c : $\mathbb{R}$)
    (h_quad : ∀ x, f x = a * x^2 + b * x + c)
    (h_vertex : f 2 = 10 ∧ ∀ x, f x ≤ f 2)  -- vertex at (2,10)
    (h_intercept1 : f 1 = 0) :               -- x-intercept at (1,0)
    f 3 = 0 := by                            -- other x-intercept at (3,0)
```

Thought: The Lean translation of the problem is appropriate because it accurately captures the intent and reasoning of the original problem. The problem's key elements, such as the vertex, axis of symmetry, and x-intercepts of the quadratic function, are correctly translated into Lean code. The logical flow of the proof mirrors the original reasoning, starting with the symmetry property of the quadratic function and using it to determine the second x-intercept. The mathematical accuracy is preserved, as the proof correctly applies the vertex property and symmetry to arrive at the solution. Furthermore, the translation aligns well with the original problem in natural language, maintaining fidelity to its structure and purpose. Overall, the translation is both faithful and complete, making it an appropriate representation of the original problem.

Judgement: Appropriate

# Example 2:

## Original natural language statement of the problem:

Draw a tangent line from the point $(4, 3)$ to the circle $(x-2)^2 + (y-1)^2 = 1$. What is the equation of the line that passes through the two tangent points?

## Translated formal statement:
```lean
theorem tangent_line_equation (x y : $\mathbb{R}$) :
  let P : $\mathbb{R}$ × $\mathbb{R}$ := (4, 3)  -- Point P
  let C : $\mathbb{R}$ × $\mathbb{R}$ := (2, 1)  -- Center of first circle
  let r : $\mathbb{R}$ := 1          -- Radius of first circle
  -- Points (x,y) satisfying both circle equations
  let on_first_circle := (x - 2)^2 + (y - 1)^2 = 1
  let on_second_circle := (x - 3)^2 + (y - 2)^2 = 2
  -- If point is on both circles
  on_first_circle ∧ on_second_circle →
  -- Then it lies on the line 2x + 2y - 7 = 0
  2*x + 2*y - 7 = 0 := by
```

Thought: The Lean translation of the problem is inappropriate because it fundamentally changes the intent of the original problem. The original problem asks to derive the equation of the tangent line passing through the intersection points of two circles, but the translation assumes the equation (2x + 2y - 7 = 0) is already given and instead asks to prove that the intersection points lie on this line. This shifts the problem from a construction task to a verification task, losing the original problem's focus on deriving the result through geometric and algebraic reasoning. Additionally, the translation omits the key reasoning step of subtracting the circle equations to derive the line equation, which is central to the original problem. As a result, the translation fails to accurately represent the problem's intent and educational value, making it an incomplete and inappropriate representation.

Judgement: Inappropriate

Example3:

## Original natural language statement of the problem:

If $a,b,c,d > 0$ and $abcd = 1$ , prove that \n\n $ \frac{1}{a+b+c +1}+ \frac{1}{b+c+d+1}+\frac{1}{c+d+a+1}+\frac{1}{d+a+b+1} \le\frac{1}{a+3} +\frac{1}{b+3} + \frac{1}{c+3} + \frac{1}{d+3} $\n\n -/

## Translated formal statement:
```lean4
theorem lean_workbook_49553 (a b c d : $\mathbb{R}$) (habc : a * b * c * d = 1) : (1 / (a + b + c + 1) + 1 / (b + c + d + 1) + 1 / (c + d + a + 1) + 1 / (d + a + b + 1)) ≤ (1 / (a + 3) + 1 / (b + 3) + 1 / (c + 3) + 1 / (d + 3))  :=  by sorry
```

Thought: The Lean translation of the problem is inappropriate because the condition $a,b,c,d>0$ is ignored in the formal statement.

Judgement: Inappropriate

Example4:

## Original natural language statement of the problem:

If $a=b=c=2$ so $\sum_{cyc}\frac{(a-1)^2}{a^2+2}=\frac{1}{2}$ . We'll prove that $\frac{1}{2}$ is the answer.

## Translated formal statement:
```lean4
theorem lean_workbook_plus_1478 (a b c : $\mathbb{R}$) (ha : a = 2) (hb : b = 2) (hc : c = 2) : (a - 1) ^ 2 / (a ^ 2 + 2) + (b - 1) ^ 2 / (b ^ 2 + 2) + (c - 1) ^ 2 / (c ^ 2 + 2) = 1 / 2   :=  by sorry
```

Thought: The Lean translation of the problem is appropriate because it accurately captures the assumptions and the goal in the natural language statement.

Judgement: Appropriate

## Original natural language statement of the problem:

{informal_statement}

## Translated formal statement:
```lean4
{formal_statement}
```

\end{plainleancode}

\section{Details for Informal-Based Scaffolded Data Synthesis}\label{sec:informal-scaffolded}

In this section, we provide all the details for informal-based scaffolded data synthesis. We start with the prompt template for different LLM queries, including the prompt template to generate harder variant, simpler/sub-problem, as well as the prompt template to judge the simplicity and correctness. Then, we provide more details for the synthesis pipeline.

\paragraph{Prompt template for solving the original problem} Note that for both input with natural language problem and statement written in Lean, we use the same prompt because we find that general-purpose models can understand the Lean although they cannot generate the whole proof correctly.

\begin{plainleancode}
Solve the following math problem probably written in Lean 4:
{problem}
Provide a detailed solution. Note that you don't need to prove the problem in Lean 4, just provide a detailed solution in natural language or math notation.
\end{plainleancode}

\paragraph{Prompt template for generating sub/simpler problems}

\begin{plainleancode}
I will give you a math problem along with its full solution.
Your task is to generate simpler problems which may enable a student to build up the skills to solve the given problem. Each simpler problem should reflect the idea of a core step in the solution.
Each generated problem must:
(1) Be completely self-contained and standalone: it should be clearly stated as an independent question that someone could read and work on without seeing the original problem and the other generated problems.
(2) Be purely proof-based: stated explicitly as a question of the form "Prove that...".
(3) Be related to the core steps in the solution of the original problem or reflect the core idea, and should not be just trivial and straightforward derivation, plug-in calculation, or solving simple equations.
After generating the problems, carefully evaluate your own output and perform the following checks:
(1) Ensure each problem is fully self-contained: check that it does not rely on undefined variables, terms, or concepts from the original problem or other problems.
(2) Ensure the set of problems is diverse, covering different steps or aspects of the original reasoning.
(3) Ensure that the problems are not too simple or trivial, and that they require a meaningful proof. Avoid trivial and straightforward derivation, plug-in calculation, or solving simple equations.
Wrap each final selected problem between the tags <newproblem> and </newproblem> to make it easy to extract.  
Do not include anything else. 
Problem: {problem}
Solution: {solution}
\end{plainleancode}

\paragraph{Prompt template for generating harder variant of problems}

\begin{plainleancode}
I now have a math problem and its solution at hand, and I would like you to modify the problem to generate a diverse set of new problems based on it. Below is the problem (probably written in Lean 4):
---
{problem}
---

Below is the solution:
---
{solution}
---

Now I would like you to generate at least 10 new math problems, each clearly different from the original. For example, you can make the following modifications to make the problem different:
Change the number in the original problem to generate a new problem.
Transform the algebraric formula such that it needs more simplification. For example, change cos(x) in the original problem into 1 - 2 sin(x/2)^2.
Add more terms in the inquality. For example, the original problem need to show that f(x) is non-negative, you transform the problem into f(x) + (1/x - a)^2 is non-negative, or f(x) \cdot (x - b)^2 is non-negative.
Lift a real variable into a complex number, a vector or even matrix. For example previously when solving quadratic equation in real space, you now change it to complex field, which might lead to slightly different solutions. Or the original problem considers planer geometry, you modify the problem into 3 dimensional geometry.
Substitute a variable into a more complex algebraic form, which includes but not limited to changeing a variable into a polynomial, exponential, logarithm, or even trigonometry. For example, you change variable x in the original problem into y^2 or even y^n in the new problem. Another example is that you are given some condition like a + b + c = 1 where a, b, c are all positive, and you need to show f(a, b, c) is non-negative. You can change the condition into x y + y z + z x = x y z with x, y, z positive, which is equivalent to 1/x + 1/y + 1/z = 1, and modify the statement to show that f(1/x, 1/y, 1/z) is non-negative.
Use the conclusion in the problem as a step to solve another problem. For example, the original problem need to solve equation f(x) = 0, where x is the variable. Now you change it to solve f( exp(x) ) = 0, f( ln(x) ) = 0, or f( cos(x) ) = 0, or even f( x^2 ) = 0.
These are just some example transformations you can try, and you are not limited to these transformations. Do not overcomplicate the problem, and it is acceptable to make simple transformations. The new math problem should not be simpler than the example problems I provided, i.e., you should not simplify the original problem and make it the new problem.
Each generated problem must:
(1) Be purely proof-based: stated explicitly as a question of the form "Prove that…". Do not generate problems that ask to "determine", "compute", "evaluate", "find", or similar.
(2) Be completely self-contained and standalone.
(3) Be mathematically valid and solvable.
(4) Be meaningfully different to ensure diversity.
Wrap each generated problem between the tags <newproblem> and </newproblem> to make it easy to extract.
Do not include anything else.
\end{plainleancode}

\paragraph{Prompt template to judge the correctness and simplicity of the synthesized statement}

\begin{plainleancode}
I will give you a math problem written in Lean 4, and I will ask you to determine (1) whether the problem is correct or not; and (2) whether the problem is too simple to prove or disprove in Lean 4. Here is a list of too simple problems in Lean 4: simple calculations, simple algebraic manipulations, solving single variable linear equations (by just a 1-step calculation), and inequalities proved by an easy sum-of-squares technique. However, do not include inequality proving with the square root since that might be more complex. Please do not label other problems, such as other more complex inequalities, limits, and integrals, as simple problems. Also, please do not label problems that deals with integers (more related to number theory), higher order roots, complex numbers, matrices, polynomials, group, finite-sum, or functional equations (e.g. Let $ f $ be a function such that $$ f(x^2) = xf(x) $$ for all $$ x \\in R. $$ Prove that $ f $ is an odd function, i.e., $$ f(-x) = -f(x) $$ for all $$ x \\in R $$.), since these problems might shed lights on other hard problems.
Please carefully analyze the problem and provide an explanation of your reasoning.
For judging the correctness, if the problem is correct, respond with "yes"; if the problem is incorrect, respond with "no"; if you are not sure, respond with "unsure".
For judging the simplicity, if the problem is too simple to prove or disprove in Lean 4, respond with "yes"; if the problem is not too simple, respond with "no"; if you are not sure, respond with "unsure". Please refer to the list of too-simple problems I provided.
Wrap your judgment between <judge> and </judge> to make it easy to extract. You should first answer yes or no to the correctness and then answer yes or no to the simplicity, separated by a comma. For example, if you think the problem is correct and not too simple, you should respond with <judge>yes, no</judge>. If you are not sure about the correctness, you can respond with <judge>unsure, no</judge> because at least this problem is not simple. If you think the problem is incorrect but it is not too easy to disprove, you can respond with <judge>no, no</judge>. If you think the problem is correct but too simple to prove or disprove, you can respond with <judge>yes, yes</judge>. Again, when judging the simplicity, please do not label inequality proving with the square root as simple, since that might be more complex than expected. Please do not include problems such as more complex inequalities, or problems that include limits, and integrals. Also, please do not label problems that deals with integers (more related to number theory), higher order roots, complex numbers, matrices, polynomials, group, finite-sum, or functional equations, as simple problems, since these notions are more likely to related to hard problems and proving problems related to these concepts in Lean 4 might not be easy even if the problem is straight-forward in natural language.
Do not include anything else. 
Problem: {formal_statement}
\end{plainleancode}

\paragraph{More details for informal-based scaffolded data synthesis} In the following part, we discuss some of the design choices and details for the scaffolded data synthesis pipeline.
\begin{enumerate}
    \item \textbf{(Informal statements generation.)} For the informal statements generation, one design choice is whether to generate multiple questions during the same inference, or to generate multiple times while only generating 1 (or very few) questions during one inference. From our experiment, we observe that for the Qwen3-32B model, generating multiple questions during the same inference is better, since the generated questions are likely to be different. Otherwise, there might be very similar questions among different generations. Besides, we also find that repeatedly generating a single problem multiple times doesn't significantly increase the number of different problems. Thus, for efficiency considerations, we only query the LLM (Qwen3-32B) once for generating multiple hard variants/simple problems of a given math problem.
    \item \textbf{(Formalization and quality checking.)} For each generated informal statement, we call the trained formalizer to formalize the problem twice. Then, we then query Qwen3-8B for 3 times for each formalization, and decide if the formalization is aligned with the informal statement using majority voting (among three queries). We decide to formalize each informal statement twice in order to balance the efficiency and the number of generated problems. We only keep at most one formalization for each generated informal statement.
    \item \textbf{(Negation and difficulty filtering.)} For each formalization, we query Qwen3-32B for 4 times, where each inference judges the correctness and the simplicity simultaneously. The final correctness is judged by strict majority voting among the 4 judges, while the final simplicity is determined if all 4 judges think the problem is easy. We use such strict criteria to minimize the probability of discarding hard and valuable problems. The efficiency for the judging is high, even if we call Qwen3-32B 4 times for each formalization, since a lot of the time Qwen3-32B enters the ``fast thinking'' mode (no long chain-of-thought for reasoning).
    \item \textbf{(Final deduplication.)} At the end of the pipeline, we filter out duplicated statements under exact match.
\end{enumerate}

\section{RL Training Details}

We further explain our RL training in detail.

\subsection{RL Implementation}
\label{app:rl_details}

\begin{figure}[!th]
    \centering
    \includegraphics[width=\linewidth]{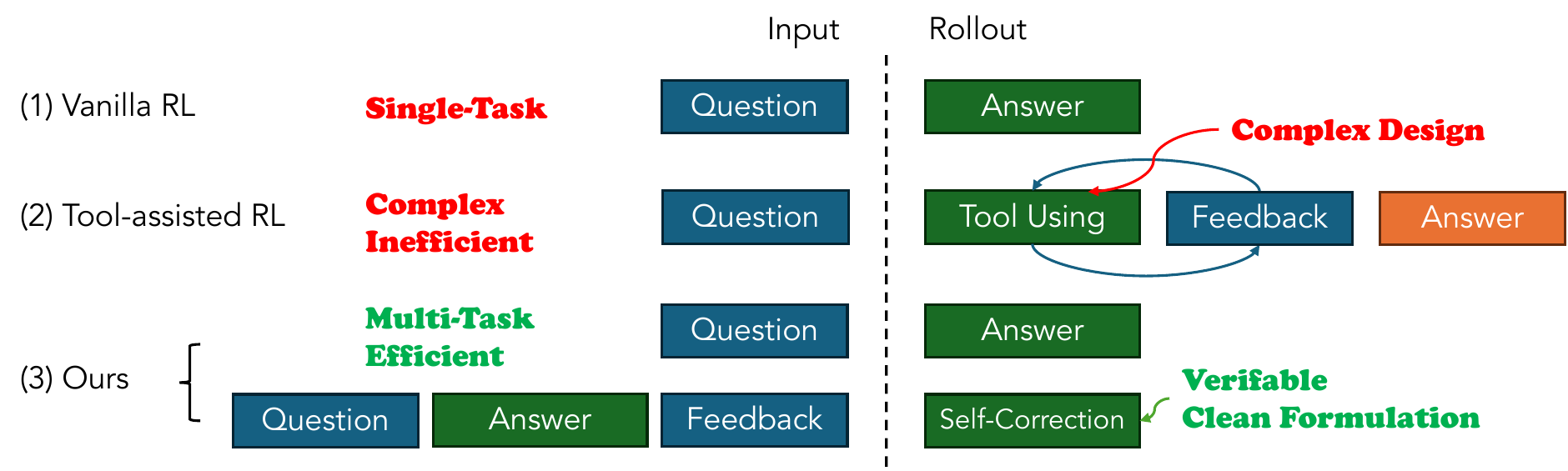}
    \caption{Illustrative figure for our multi-task RL. 
    This pipeline improves models' performance on both the whole proof and the self-correction at the same time without additional design on the framework or algorithm. }
    \label{fig:rl}
\end{figure}

We begin by collecting 50K challenging statements and 50K self‑correction samples. Each self‑correction sample contains a statement, a output generated by the SFT model, and an associated error message. During RL training, we consumed approximately 46K and 64K unique inputs for the 8B and 32B models on 1 epoch, respectively. 
We use the VeRL framework~\citep{sheng2024hybridflow} with several key setups: 
We adopt a batch size of 128 and n of 8 for parallel rollouts and reward function calls (via the Lean compiler), while using a mini-batch size of 32, accepting a certain degree of off-policy training in exchange for a higher frequency of policy optimization.
For the dynamic sampling, we set the over-sample batch size equal to three times of train batch size and filter out inputs with a pass rate equal to 0 or higher than 0.75. 
We use a maximum prompt length of 16K for those long inputs in the self-correction task.
We set the maximum response length to 24K to support a reasonably sized reasoning trajectory while staying within the Qwen3 model’s native 40K context window, ensuring generation quality.
We enable the overlong penalty with a 4K overlong buffer and set the overlong penalty factor to 1. 
We use token-averaged policy loss and do not use KL divergence or entropy terms in our training objective.

\begin{figure}[htbp]
    \centering
    \begin{subfigure}[b]{0.48\textwidth}
        \includegraphics[width=\textwidth]{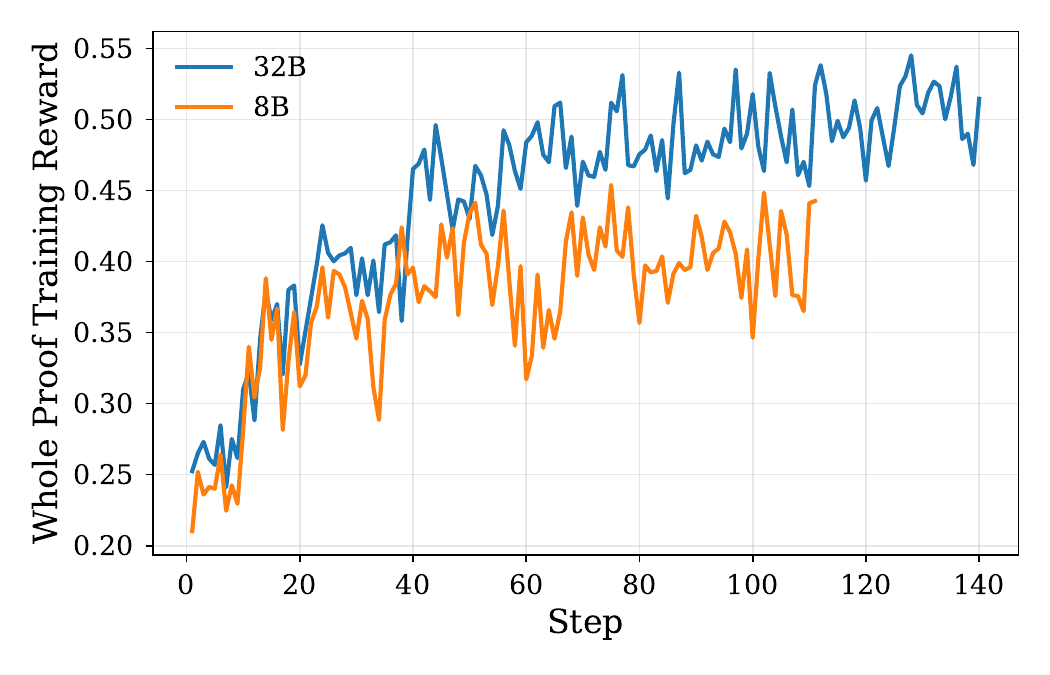}
        \label{fig:completion_reward}
    \end{subfigure}
    \hfill
    \begin{subfigure}[b]{0.48\textwidth}
        \includegraphics[width=\textwidth]{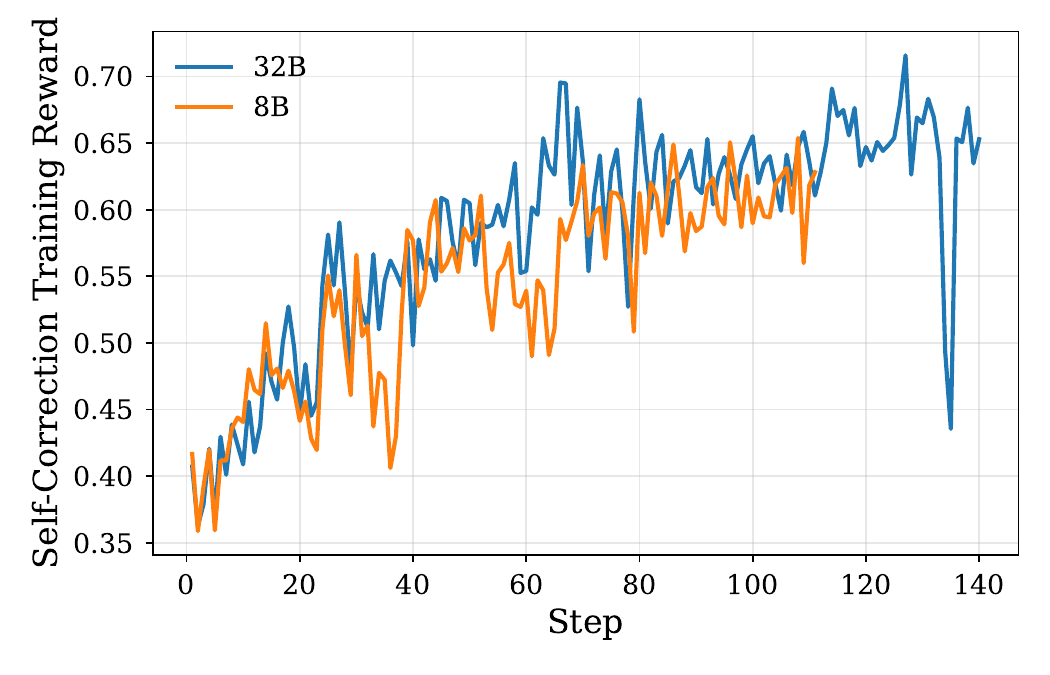}
        \label{fig:revision_reward}
    \end{subfigure}
    
    \vspace{-0.3cm}
    
    \begin{subfigure}[b]{0.48\textwidth}
        \includegraphics[width=\textwidth]{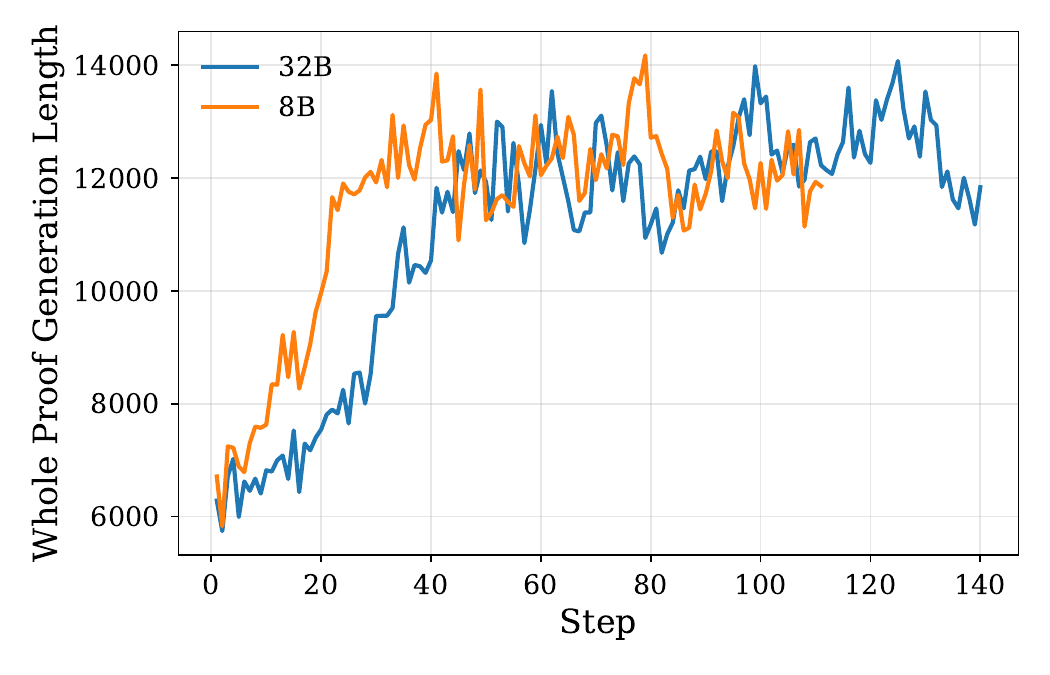}
        \label{fig:completion_length}
    \end{subfigure}
    \hfill
    \begin{subfigure}[b]{0.48\textwidth}
        \includegraphics[width=\textwidth]{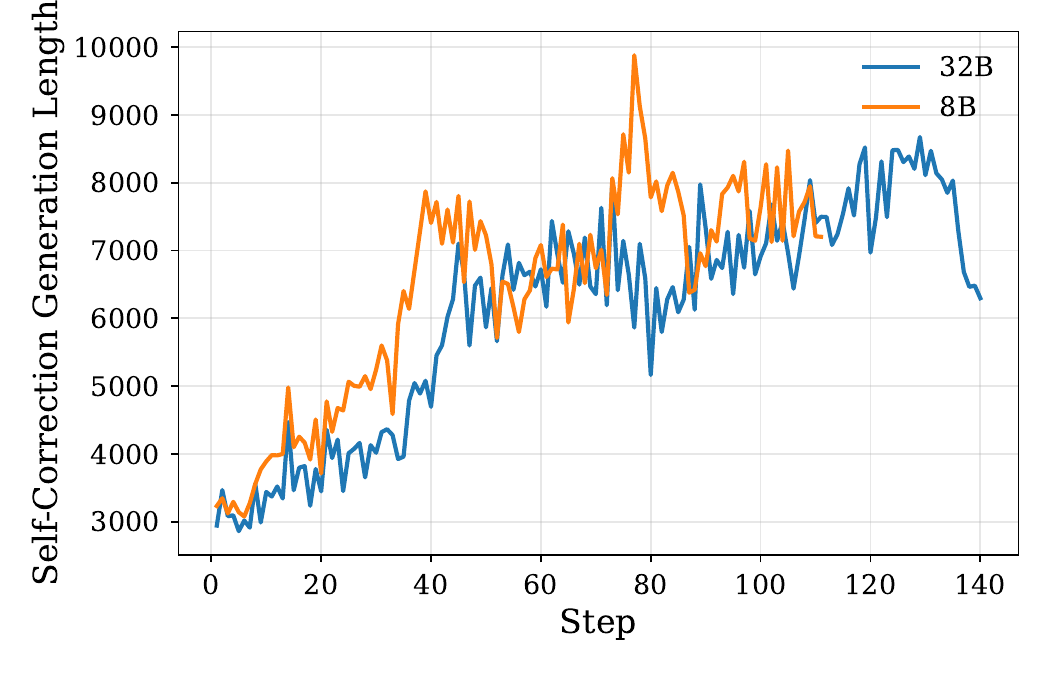}
        \label{fig:revision_length}
    \end{subfigure}
    \vspace{-0.5cm}
    
    \caption{Training curves comparing 32B and 8B models: training rewards (top row) and generation lengths (bottom row) for whole proof completion and self-correction tasks.
    Note that the rewards are averaged over all generated rollouts, including those then filtered by the dynamic sampling strategy, thus reflecting the policy improvement.
    }
    \label{fig:rl_per_source_training_curves}
\end{figure}

\subsection{Further Discussion on RL}

For the design of training with self-correction, we explored both tool-use and multi-turn reinforcement learning. 
However, the former not only requires a dedicated tool-calling design, but also demands strong model capability to follow the tool-calling protocol, especially in complex and lengthy formal language scenarios like Lean, which is particularly challenging for our relatively small model. 
Meanwhile, multi-turn approaches introduce various engineering challenges, especially on the rollout engine side, such as asynchronous generation. 
Moreover, on the algorithm side, the effectiveness and efficiency of multi-turn RL still require further validation. 
We have explored some preliminary approaches, but they remain immature.
Therefore, we adopt a more straightforward implementation that is natively integrated into the current RL framework, as illustrated in Figure~\ref{fig:rl}.
In this setup, we have two different type of inputs for RL and detailed training curves are shown in Figure~\ref{fig:rl_per_source_training_curves}.
Furthermore, for the algorithmic design, like the advantage estimator, we explored different popular GRPO variants but did not observe significant differences.